\documentclass[preprint,12pt]{elsarticle}
\usepackage[pdfborder={0 0 0}]{hyperref}
\usepackage{amsthm,amsmath,amsfonts,latexsym,amssymb}
\usepackage{graphics,fancyhdr,graphicx}
\usepackage[ruled,linesnumbered]{algorithm2e}
\usepackage[margin=1in]{geometry}
\usepackage{multirow,booktabs}
\usepackage[table]{xcolor}
\usepackage{listings}
\usepackage{tabularx}
\usepackage{placeins}
\usepackage{comment}
\usepackage{lscape}
\usepackage{bm}
\usepackage{hyperref}
\usepackage{mathrsfs}
\usepackage{arydshln}
\usepackage{appendix}
\definecolor{custom-blue}{RGB}{3,69,173}
\hypersetup{
    colorlinks = true,
    urlcolor   = blue,
    citecolor  = custom-blue,
}
\biboptions{numbers,sort&compress}
\definecolor{listinggray}{gray}{0.9}
\definecolor{lbcolor}{rgb}{0.9,0.9,0.9}
\definecolor{Darkgreen}{RGB}{0,100,0}
\usepackage{caption}

\usepackage{float}
\usepackage{booktabs}
\usepackage{array}
\usepackage{multirow}
\usepackage{tabularray}
\usepackage{algpseudocode}
\usepackage{mathtools}
\usepackage{subcaption}
\usepackage{enumitem}

\makeatletter
\def\ps@pprintTitle{%
 \let\@oddhead\@empty
 \let\@evenhead\@empty
 \def\@oddfoot{}%
 \let\@evenfoot\@oddfoot}
\makeatother

\begin{document}
\abovedisplayskip=6.0pt
\belowdisplayskip=6.0pt


\newcommand{\SG}[2][blue]{{\textcolor{#1}{#2}}}

\begin{frontmatter}

\title{Synergistic Learning with Multi-Task DeepONet for Efficient PDE Problem Solving}

\author[1]{Varun Kumar}
\ead{varun_kumar2@brown.edu}
\author[2]{Somdatta Goswami\corref{cor1}}
\ead{sgoswam4@jhu.edu}
\author[2]{Katiana Kontolati}
\ead{katiana.kontolati@gmail.com}
\author[2]{Michael D. Shields}
\ead{michael.shields@jhu.edu}
\author[1,3]{George Em Karniadakis}
\ead{george_karniadakis@brown.edu}

\address[1]{School of Engineering, Brown University}
\address[2]{Department of Civil and Systems Engineering, Johns Hopkins University}
\address[3]{Division of Applied Mathematics, Brown University}

\cortext[cor1]{Corresponding author.}

\begin{abstract}
\noindent
Multi-task learning (MTL) is an inductive transfer mechanism designed to leverage useful information from multiple tasks to improve generalization performance compared to single-task learning. It has been extensively explored in traditional machine learning to address issues such as data sparsity and overfitting in neural networks. In this work, we apply MTL to problems in science and engineering governed by partial differential equations (PDEs). However, implementing MTL in this context is complex, as it requires task-specific modifications to accommodate various scenarios representing different physical processes. To this end, we present a multi-task deep operator network (MT-DeepONet) to learn solutions across various functional forms of source terms in a PDE and multiple geometries in a single concurrent training session. We introduce modifications in the branch network of the vanilla DeepONet to account for various functional forms of a parameterized coefficient in a PDE. Additionally, we handle parameterized geometries by introducing a binary mask in the branch network and incorporating it into the loss term to improve convergence and generalization to new geometry tasks. Our approach is demonstrated on three benchmark problems: (1) learning different functional forms of the source term in the Fisher equation; (2) learning multiple geometries in a 2D Darcy Flow problem and showcasing better transfer learning capabilities to new geometries; and (3) learning 3D parameterized geometries for a heat transfer problem and demonstrate the ability to predict on new but similar geometries.  Our MT-DeepONet framework offers a novel approach to solving PDE problems in engineering and science under a unified umbrella based on synergistic learning that reduces the overall training cost for neural operators.
 \end{abstract}
\end{frontmatter}

\section{Introduction}
\label{sec:intro}

In scientific machine learning, we can solve partial differential equations (PDEs) by finding the solution operator, known as the neural operator (NO). The NO takes different functions as inputs, such as initial and boundary conditions, and maps them to the solution of the PDE. Traditional numerical methods such as finite difference, finite element, and spectral methods are generally used to compute solutions to PDEs. There is an increasing interest in using scientific machine learning methods to solve PDEs in real time across diverse applications. However, these real-time methods can be computationally expensive when dealing with high dimensional PDEs, and incorporating experimental measurement data as model inputs is often not possible. Additionally, the solution must be recomputed for minor changes in the input function or the geometry domain that add to the computational burden for the users. 

Recently, Deep neural networks (DNNs) have been employed in NOs  \cite{lu2022comprehensive,goswami2022physics2} to approximate mappings between infinite-dimensional Banach spaces, in contrast to the finite-dimensional vector space mapping learned through functional regression in conventional DNNs. Frameworks such as deep operator network (DeepONet) \cite{lu2021learning} and integral operators, which include architectures like the Fourier neural operator (FNO) \cite{li2020fourier}, the wavelet neural operator (WNO) \cite{tripura2022wavelet}, the Laplace neural operator (LNO) \cite{cao2023lno}, and convolutional neural operator (CNO) \cite{raonic2023convolutional}, have demonstrated significant potential over a range of applications. While the early success of NOs has been promising, their predictive performance is often limited by the availability of labeled data for training. Collecting large labeled datasets for each task can be computationally intractable, especially for high-fidelity or multi-scale models. Multi-task learning (MTL) is an alternative mechanism aimed at leveraging useful information from related learning tasks to address data sparsity and overfitting issues \cite{caruana1997multitask}. This inductive transfer mechanism trains tasks in parallel while using a shared representation, assuming that the tasks are associated with each other and that shared information among them can lead to synergistic learning performance.

MTL has been explored in traditional machine learning tasks such as natural language processing, computer vision, and healthcare to improve generalization scenarios with limited training data. There are two prevalent techniques for using MTL based on the connections between the learning tasks: hard parameter sharing and soft parameter sharing. Hard parameter sharing uses a common hidden layer for all tasks, while soft parameter sharing regularizes the distance between parameters in different models. Hard parameter sharing techniques are useful when tasks have different input data distributions but similar output conditional distributions (\textit{i.e.}, $P(\mathbf{x}_s) \ne P(\mathbf{x}_t)$ and $P(\mathbf{y}_s\vert\mathbf{x}_s) = P(\mathbf{y}_t\vert\mathbf{x}_t)$), typically referred to as covariate shift. MTL has received significant attention in the domain of computer vision. Some notable works include Liu et al.'s \cite{liu2016_MTLreview} deep fusion with LSTM modules, and Long et al.'s \cite{long2017_MTLreview} joint adaptation networks for transfer learning. In the computer vision, MTL methods such as PAD-Net \cite{padnet_MTLreview}, MTAN \cite{liu2019end_MTLreview}, and cross-stitch networks \cite{misra2016_MTLreview} have achieved significant advancements in tasks such as depth estimation, scene parsing, and surface normal prediction. Recently, Reed et al. \cite{reed2022_GATO} introduced GATO, a generalist agent using the transformer architecture to handle multiple tasks like image captioning, gaming, playing Atari, \textit{etc.} simultaneously, demonstrating remarkable versatility. Liu \cite{yang2023context} introduced an in-context learning paradigm to learn a common operator mapping from a set of differential equations. Liu et al. utilize a transformer framework where key-value pairs are used as input queries for predicting the output solution of a differential equation. The key-value pair represents conditions that define the differential equation such as the initial condition in a temporal problem. This framework shows good generalization capabilities due to its in-context learning paradigm. 

\begin{figure}[h]
\begin{center}
\includegraphics[width=\textwidth]{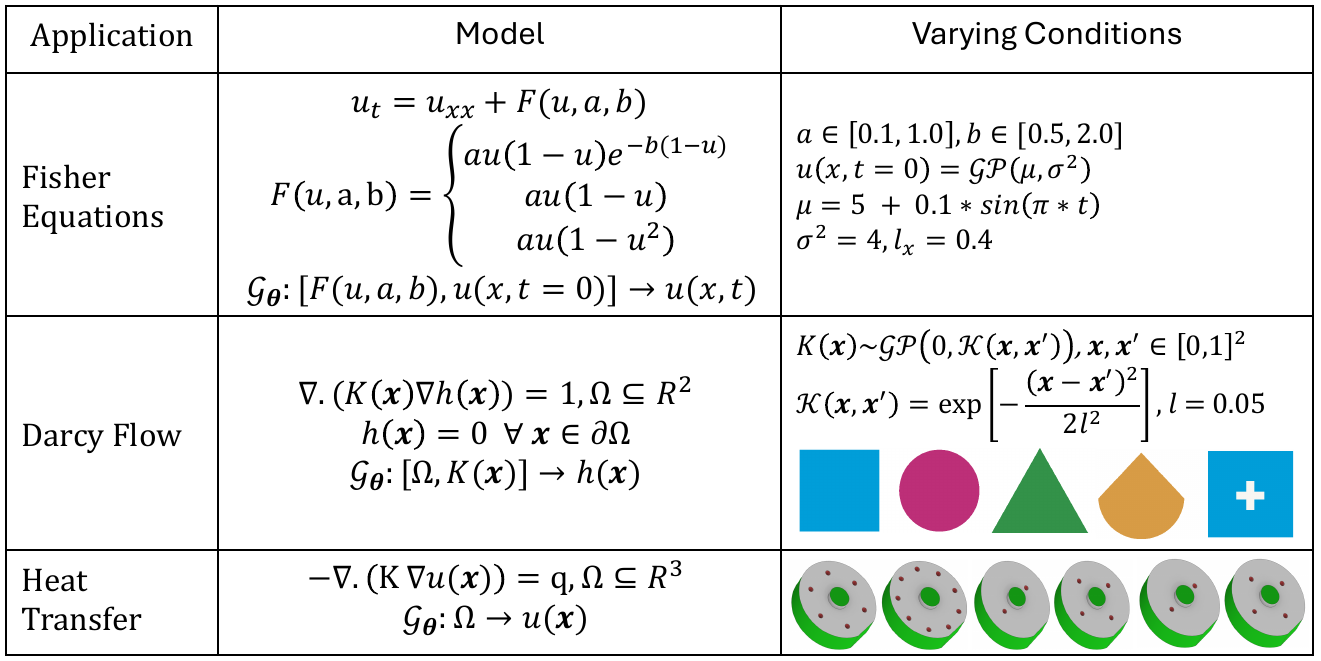}
\caption{A schematic representation of the operator learning benchmarks and MTL scenarios considered in this study.}
\label{fig:applications}
\end{center}
\captionsetup{justification=centering}
\end{figure}

Liu's research presents a good opportunity for developing multi-task operator frameworks that can be applied to realistic problems, particularly in engineering and life sciences. One significant challenge for the operator network relates to handling varying geometric domains, a problem that is not addressed in current frameworks including Liu's. In science and engineering tasks, the application of MTL frameworks is complicated, since PDEs with the same initial or boundary conditions can represent vastly different physical systems. However, a group of tasks can share the same marginal distribution of inputs or even the same input function, while their conditional output distributions may differ significantly. This scenario, known as conditional shift, occurs when $P(\mathbf{x}_s) = P(\mathbf{x}_t)$ and $P(\mathbf{y}_s\vert\mathbf{x}_s) \ne P(\mathbf{y}_t\vert\mathbf{x}_t)$. In such cases, transfer learning, often referred to as soft parameter sharing, has shown success. In our recent study \cite{goswami2022transfer_learning}, we proposed the idea of transfer learning within the DeepONet (TL-DeepONet) architecture to enable the knowledge transfer from one task to a related but different task, allowing task-specific learning under conditional shift. However, we found limitations in this framework when attempting to transfer knowledge across varying geometries, such as changes in internal and external boundaries of the target geometry. In this work, we aim to extend DeepONet's capability to train multiple parameterized PDEs on multiple domains concurrently, enhancing the generalizability of a single network and thereby improving the transfer learning process to new geometric domains. To achieve this, we introduce multi-task DeepONet (MT-DeepONet) designed to predict solutions for different but correlated tasks in a single training session. Figure \ref{fig:applications} illustrates the problem statements for the MTL problems considered in this study of MT-DeepONet. The main contributions of this work are summarized as follows:\vspace{-6pt}

\begin{itemize}
    \item \textbf{Extension of DeepONet for concurrent training over multiple tasks}: We develop MT-DeepONet to approximate the solution for multiple tasks (different parametric conditions and source terms) simultaneously, without requiring re-training.\vspace{-8pt}
    \item \textbf{Improved generalizability}: We investigate the generalization ability of MT-DeepONet for knowledge sharing across different geometries as an extension to \cite{goswami2022transfer_learning}. We demonstrate improvement in target model learning across varied geometries using the MTL source model as compared to a single-task source model.\vspace{-8pt}
    \item \textbf{Enhanced knowledge transfer}: We introduce a masking operation that enables our MT-DeepONet to learn solutions across varied geometries. Our methodology is demonstrated by learning solutions for $2$D Darcy flow equations across multiple geometric domains and steady-state heat transfer in multiple $3$D plate designs parameterized by the location and number of heating sources. \vspace{-8pt}
\end{itemize}

The paper is organized as follows. In Section \ref{sec:MTL}, we provide a brief review of the original DeepONet framework followed by a description of the proposed MT-DeepONet framework. In Section \ref{sec: prob_def}, we present a comprehensive collection of problems for which the proposed MT-DeepONet has been extensively studied. We discuss the data generation process and include results and comparisons for multiple examples. Finally, we summarize our observations and provide concluding remarks in Section \ref{sec:summary} along with some limitations of the framework.

\section{Multi-task learning in neural operators}
\label{sec:MTL}

Neural operators learn nonlinear mappings between functional spaces on bounded domains, offering a unique framework for real-time solution inference for complex parametric PDEs. Here, `parametric PDEs' refer to PDE systems with parameters that vary over a certain range. Typically, DeepONet (our choice of operator network for this work) is trained on a fixed domain, $\Omega$, for varying parametric conditions drawn from a distribution. In this work, we introduce MT-DeepONet, which enables concurrent training of multiple functions (leading to different dynamics) and multiple geometries, along with varying parametric conditions. This section provides a brief overview of the DeepONet architecture and extends it to discuss our multi-task DeepONet framework.

\subsection{Deep operator network}
\label{subsec:neural_operators}

The goal of operator learning is to learn a mapping between two infinite-dimensional spaces on a bounded open set $\Omega \subset \mathbb{R}^D$, given a finite number of input-output pairs. Let $\mathcal{U}$ and $\mathcal{S}$ be Banach spaces of vector-valued functions defined as:
\begin{align}
    &\mathcal{U} = \{\Omega; u: \mathcal{X} \to \mathbb{R}^{d_u}\}, \quad \mathcal{X}\subseteq \mathbb{R}^{d_x}\\
    &\mathcal{S} = \{\Omega; s: \mathcal{Y} \to \mathbb{R}^{d_s}\}, \quad \mathcal{Y}\subseteq \mathbb{R}^{d_y}, 
\end{align}
where $\mathcal{U}$ and $\mathcal{S}$ denote the set of input functions and the corresponding output functions, respectively. The operator learning task is defined as $\mathcal{G}: \mathcal{U} \to \mathcal{S}$. The objective is to approximate the nonlinear operator, $\mathcal{G}$,  via the following parametric mapping:
\begin{equation}
    \mathcal{G}: \mathcal{U} \times \mathbf{\Theta} \rightarrow \mathcal{S} \quad \text{or} \quad \mathcal{G}_{\boldsymbol{\theta}}: \mathcal{U} \rightarrow \mathcal{S}, \quad \boldsymbol{\theta} \in \mathbf{\Theta},
\end{equation}
where $\mathbf{\Theta}$ is a finite-dimensional parameter space. In the standard setting, the optimal parameters $\boldsymbol{\theta}^*$ are learned by training the neural operator with a set of labeled observations $\mathcal{D} = \left\{(u^{(i)}, s^{(i)})\right\}_{i=1}^N$, which contains $N$ pairs of input and output functions. When a physical system is described by PDEs, it involves multiple functions, such as the PDE solution, the forcing term, the initial condition, and the boundary conditions. We are typically interested in predicting one of these functions, which is the output of the solution operator (defined on the space $\mathcal{S}$), based on the varied forms of the other functions, i.e., the input functions in the space $\mathcal{U}$.

The deep operator network (DeepONet) is inspired by the universal approximation theorem for operators \cite{chen1995universal}. The architecture of DeepONet comprises two deep neural networks: the branch network and the trunk network. The branch network encodes the input functions $\mathcal{U}$ at fixed sensor points $\{x_1, x_2, \dots, x_m\}$, while the trunk network encodes the information related to the spatio-temporal coordinates $\zeta = \{x_i, y_i, t_i\}$ where the solution operator is evaluated. The trunk network takes these spatial and temporal coordinates to compute the loss function. The solution operator for an input realization $u_1$ can be expressed as:
\begin{equation}\label{eq:output_deeponets}
    \begin{split}
      \mathcal G_{\boldsymbol \theta}(u_1)(\zeta) &= \sum_{i = 1}^p b_i \cdot tr_i = \sum_{i = 1}^{p}b_i(u_{1}(x_1), u_{1}(x_2), \ldots, u_{1}(x_m))\cdot tr_i(\zeta),  
\end{split}
\end{equation}
where $\{b_1, b_2, \ldots, b_p\}$ are the output embeddings of the branch network and $\{tr_1, tr_2, \ldots, tr_p\}$ are the output embeddings of the trunk network. In Eq.~\eqref{eq:output_deeponets}, $\boldsymbol{\theta} = \left(\mathbf{W}, \mathbf{b} \right)$ represents the trainable parameters of the network including weights, $\mathbf{W}$, and biases, $\mathbf{b}$. The optimized parameters $\boldsymbol{\theta}^*$, are obtained by minimizing a standard loss function ($\mathcal{L}_1$ or $\mathcal{L}_2$) using a standard optimization algorithm. 

The DeepONet model provides a flexible framework that allows the branch and trunk networks to be configured with different architectures. For equispaced discretization of the input function, a convolutional neural network (CNN) can be utilized for the branch network architecture, whereas a multilayer perceptron (MLP) is often employed for a sparse representation of the input function. An MLP is commonly used for the trunk network to handle the low-dimensional evaluation points, $\zeta$. Since its inception, standard DeepONet has been applied to address complex, high-dimensional systems \cite{di2021deeponet,kontolati2022influence,goswami2022physics,oommen2022learning,cao2023deep,taccari2024developing,borrel2024sound}. Recent extensions for DeepONet have explored multi-fidelity learning \cite{de2022bi,lu2022multifidelity,howard2022multifidelity}, integration of multiple-input continuous operators \cite{jin2022mionet,goswami2022neural}, hybrid transferable numerical solvers \cite{zhang2022hybrid,kahana2023geometry}, resolution independent learning \cite{bahmani2024resolution}, transfer learning \cite{goswami2022deep}, physics-informed learning to satisfy the underlying PDE \cite{wang2021learning,goswami2022physics2,mandl2024separable}, and learning in latent spaces \cite{kontolati2024learning}. 

\subsection{Multi-task deep operator network (MT-DeepONet)}
\label{sec:mtdeeponet}
\noindent Neural operators are inherently data-driven models that require substantial datasets to develop a generalized solution operator for parameterized PDEs. In general, applications using DeepONet to learn the solution operator have focused on single-domain geometries, parameterizing either the source term or the initial condition with a Gaussian random field. In this work, our goal is to develop a generalized solution operator capable of accommodating various functional forms of source terms and their parameterization, across multiple geometries. The different applications explored in this study are illustrated in Figure \ref{fig:applications}. The MT-DeepONet framework is designed to: \vspace{-0.1in}
\begin{itemize}
\item Learn multiple source terms representing different physical systems in a single training process, demonstrated through the Fisher equation. \vspace{-0.1in}
\item Simultaneously learn the solution operator on different geometries, thereby improving the source model's ability to transfer knowledge to a target model. This is illustrated by solving the Darcy flow problem in various $2$D geometries and learning the temperature distribution across multiple unique $3$D engineering geometries.\vspace{-0.1in}
\end{itemize}

The primary modification in the MT-DeepONet framework occurs in the branch network of the standard DeepONet architecture. Each problem is addressed uniquely. For example, in the case of multiple source terms in the Fisher equation (see Figure \ref{fig:applications}), the source terms are represented as a polynomial to create a unique representation for each equation:

\begin{align}\label{eqn:Fisher_poly}
    F(u) = \alpha u + \beta u^2 + \gamma u^3 + \delta \mathcal{O}(u^4). 
\end{align}

\begin{figure}[ht!]
\begin{center}
\includegraphics[width=0.95\textwidth]{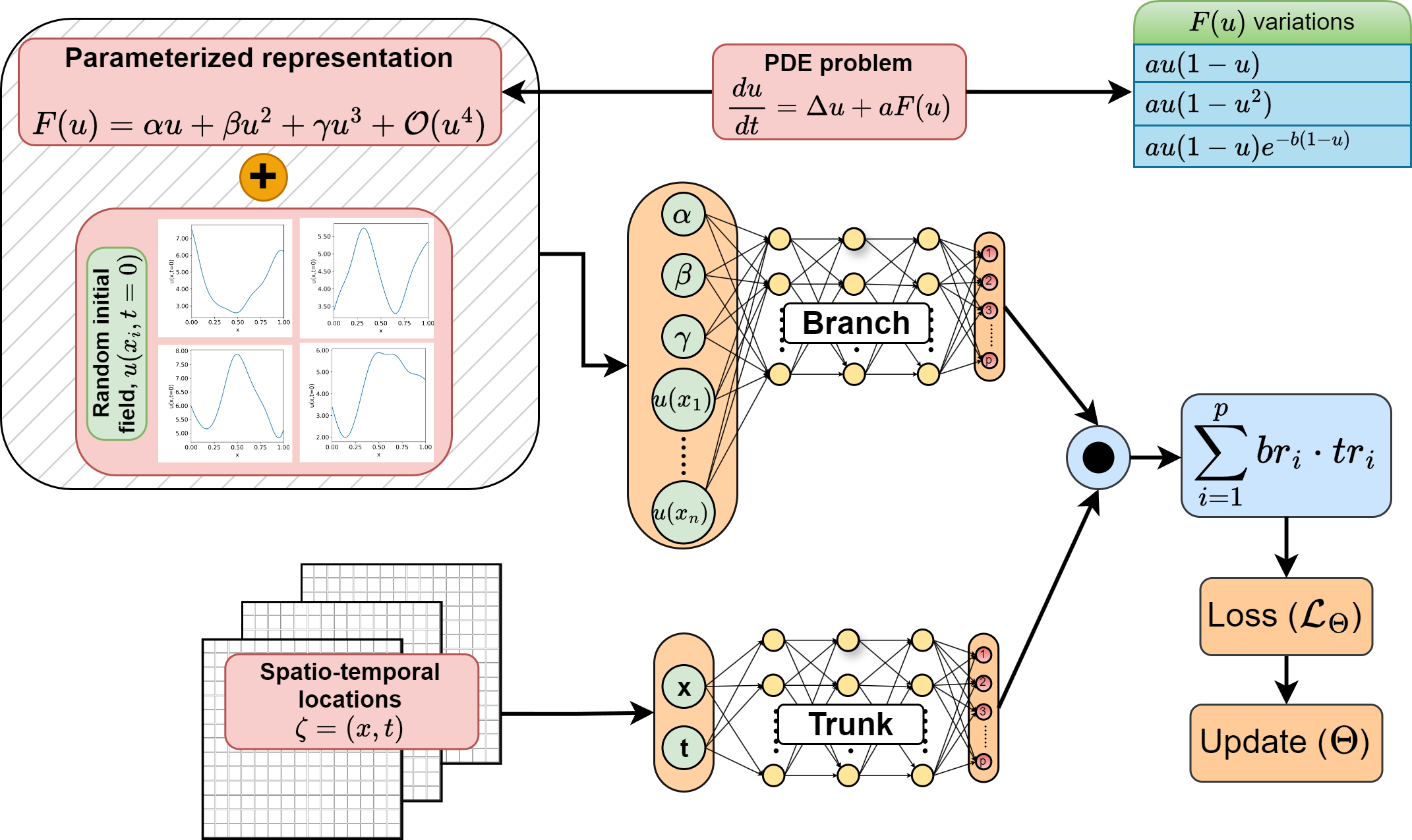}
\caption{Schematic of the MT-DeepONet designed to learn a family of parametric PDEs (Fisher equations) defined by different forcing functions, $F(u)$. The parametric representation of this equation family, along with random initial condition fields, is input to the branch network. Spatio-temporal points are input to the trunk network. The multi-task operator network learns to predict the solution field, $u$ across this parameterized family of equations and random initial solutions concurrently.}
\label{fig:schematic1}
\end{center}
\captionsetup{justification=centering}
\end{figure}

The coefficients of the dependent variable $u$ in this polynomial expression are used as inputs to the branch network, along with the random initial conditions that define the problem. A schematic of the framework is shown in Figure \ref{fig:schematic1} for understanding. Further details of the problem are discussed in Section \ref{subsec:example2}.

\begin{figure}[h]
\begin{center}
\includegraphics[width=0.75\textwidth]{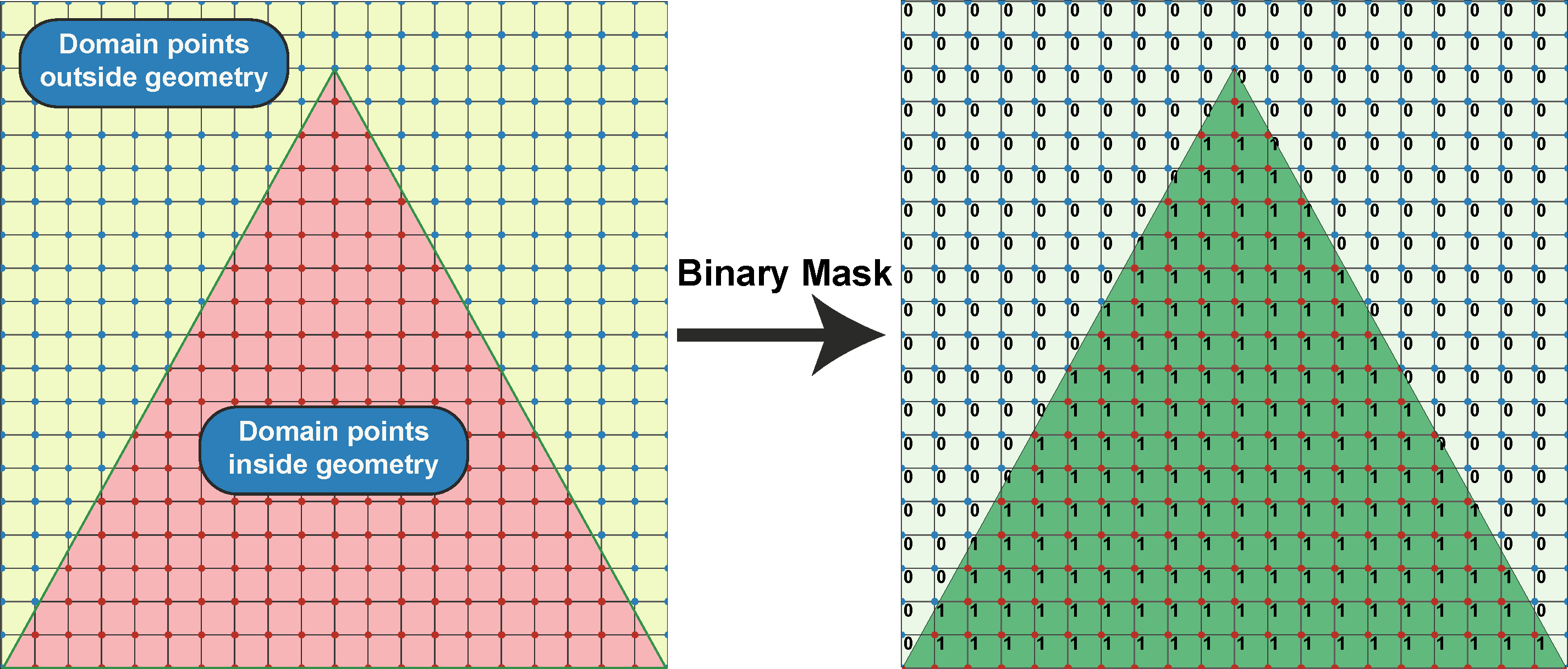}
\caption{Schematic showing the binary masking function for a triangular geometry. A uniform grid $100 \times 100$ is sampled in $\Omega \in [0,1] \times [0,1]$ and used for generating the basis function in the trunk network. The binary mask is constructed by delineating the boundaries of the domain. Grid points within the domain boundary are denoted by $1$, while those outside are denoted by $0$. The binary mask is applied to the solution from the operator network, enforcing the solution outside the geometry to be $0$, thereby aiding with network convergence.}
\label{fig:masking}
\end{center}
\captionsetup{justification=centering}
\end{figure}

To accommodate varying geometries concurrently in a single training process, we use a binary mask (array of $0$'s and $1$'s). This mask is constructed by fitting the geometry within a unit square plate for $2$D problems and a box for $3$D problems. The masking function assigns a value of $1$ to points within the boundary of the desired geometry and $0$ to points outside the boundary but within the plate or box. Figure \ref{fig:masking} illustrates the masking function with a triangular geometry within a square plate. The solution operator is defined as the product of $\mathcal{G}_{\boldsymbol{\theta}}$ (as described in Equation \ref{eq:output_deeponets}) and the binary masking function, ensuring the solution is confined within the geometry's bounds. Algorithm \ref{alg:masking_algo} details the steps for training the MT-DeepONet with a binary mask to address problems across multiple geometries. This approach also tests our hypothesis that learning multiple geometries improves the source model's ability to transfer knowledge to target models with different geometries (Section \ref{subsec:darcy}). However, given the complexity of such knowledge transfer in the diverse range of PDEs representing different physical systems, we demonstrate the effectiveness of the masking framework for learning the solution operator across unseen 3D geometries in a steady-state heat transfer problem (Section \ref{subsec:example4}).

\begin{algorithm}
\caption{Operator learning with binary mask}
\label{alg:masking_algo}
Prepare the binary mask: $\boldsymbol{M}_{binary}$\\
Input: $\boldsymbol{K}(\boldsymbol{x}), \boldsymbol{M_{binary}}$ \\
Output: $\mathcal{G}(\boldsymbol{x}, \boldsymbol{K}, \boldsymbol{M})$ \\
Branch and Trunk network parameters: {$\boldsymbol{\Theta}_{br}$,  $\boldsymbol{\Theta}_{tr}$} \\
Number of Epochs: n\\
\For{$n \leq n_{max}$}{
$br_{k} \leftarrow \mathcal{F}(\boldsymbol{K}(\boldsymbol{x}), \boldsymbol{M}, \boldsymbol{\Theta}_{br})$\\
$tr_{k} \leftarrow \mathcal{T}(\boldsymbol{x}, \boldsymbol{\Theta}_{tr})$\\
$\mathcal{G}(\boldsymbol{x}, \boldsymbol{K}, \boldsymbol{M}, \boldsymbol{\Theta}_{br}, \boldsymbol{\Theta}_{tr}) \leftarrow \sum_{k=1}^p br_{k} \cdot tr_{k}$\\
$\mathcal{G}(\boldsymbol{x}, \boldsymbol{K}, \boldsymbol{M}, \boldsymbol{\Theta}_{br}, \boldsymbol{\Theta}_{tr}) \leftarrow \mathcal{G}(\boldsymbol{x}, \boldsymbol{K}, \boldsymbol{M}, \boldsymbol{\Theta}_{br}, \boldsymbol{\Theta}_{tr}) \cdot \boldsymbol{M}_{binary}$\\
 $\boldsymbol{\Theta}_{br}, \boldsymbol{\Theta}_{tr} \leftarrow \textbf{backprop update}$\\
 }
\end{algorithm}

MTL serves as an inductive transfer mechanism designed to enhance generalization performance compared to single-task learning. It achieves this by leveraging valuable information from multiple learning tasks and utilizing domain-specific insights embedded within training samples across related tasks. In our study, we demonstrate the effectiveness of MTL in learning solutions across diverse sets of PDEs, initial conditions, and geometries simultaneously.

\section{Numerical examples} 
\label{sec: prob_def}

In this section, we explore the capabilities of the proposed MT-DeepONet framework on three problems shown in Figure \ref{fig:applications}. Detailed information on data generation for each problem can be found in Supplementary \ref{sec:data_generation}, while specifics about the network architecture are provided in Supplementary \ref{sec:network_details}.

\subsection{Fisher Equations}
\label{subsec:example2}

The first example considers the Fisher equation proposed by Ronald Fisher in $1937$, which provides a mathematical framework for analyzing population dynamics and chemical wave propagation with diffusion \cite{fisher1937wave}. The original reaction-diffusion equation is defined as:
\begin{align}
    u_t = D u_{xx} + ru(1-u),
\end{align}
where $u$ represents population density that varies spatially and temporally, $D$ and $r$ are scalar parameters denoting the diffusion coefficient and the intrinsic growth rate, respectively. In dimensionless form, this equation is written as:
\begin{align}
    u_t = u_{xx} + u(1-u).
\end{align}
Kolmogorov, Petrovsky, and Piskunov introduced a more general form, the Fisher-KPP equation \cite{tikhomirov1991}:
\begin{align}
    u_t = D (u_{xx} + u_{yy}) + F(u), \label{eqn:Fisher_KPP_2D}
\end{align}
where the population density, $u$, varies along two spatial dimensions $(x,y)$, and the reaction term $F(u)$ must satisfy the following criteria:
\begin{subequations}
    \begin{align}
        & F(0) = F(1) = 1, \\
        & F(u) > 0, \;\; u \in (0,1), \\
        & F'(0) = \alpha, \;\; \alpha > 0, \\
        & F'(u) < \alpha, \;\; u \in (0,1).
    \end{align}
\end{subequations}
Assuming that the density $u$ is invariant along the $y$-axis, the Fisher-KPP equation in dimensionless form is re-written as:
\begin{align}\label{eqn:Fisher_KPP}
    u_t = u_{xx} + F(u), \;\; 0 \leq u \leq 1. 
\end{align}

\begin{table}[h]
\centering
\caption{Details of reaction term $F(u)$ from literature and their modified forms used in this study \cite{zhang2024discovering}.}
\label{tab:Fisher-KPP}
\begin{tabular}{@{}lccc@{}}
\toprule
\textbf{Equation} & \begin{tabular}[c]{@{}l@{}}\textbf{General Form}\\ \textbf{$\;\;\;\;\;\;\;\;F(u)$}\end{tabular} & \begin{tabular}[c]{@{}l@{}}\textbf{Our Form}\\ \textbf{$\;\;\;\;\;F(u)$}\end{tabular} & \begin{tabular}[c]{@{}l@{}}\textbf{Parameter}\\ \textbf{\;\;\;\;Range}\end{tabular} \\ \midrule
Fisher \cite{fisher1937wave} & $au(1-u)$ & $au(1-u)$ & $a \in [0.1, 1.0]$ \\ 
Newell–Whitehead–Segel \cite{patade2015approximate} & $au-bu^{q}, q \in \mathbb{N}$ & $au(1-u^{2})$ & $a \in [0.1, 1.0]$ \\ 
Zeldovich–Frank–Kamenetskii \cite{zeldovich1980flame}
& $au(1-u)e^{-b(1-u)}$ & $au(1-u)e^{-b(1-u)}$ & \begin{tabular}[c]{@{}l@{}}$a \in [0.1, 1.0]$\\ $b \in [0.5, 2.0]$\end{tabular}\\
\bottomrule
\end{tabular}
\end{table}
In the conventional operator learning task, the focus is typically on analyzing the change in density profile $u(x,t)$ over a one-dimensional spatiotemporal domain $x \in [0, 1]$ and $t \in [0, 1]$ for parameterized initial condition $u(x, t=0)$ drawn from a distribution. Due to diffusion, regions with higher density expand over time towards areas with lower density, based on the initial distribution. For our MT-DeepONet, we aim to evaluate the Fisher-KPP model to learn the density variation $u(x,t)$ over time for two tasks: ($i$) varying initial conditions of density $u(x, t=0)$, and (ii) three reaction functions $F(u)$ in a single training cycle, considering different functional of $F(u)$ that are separately parameterized by two scalar coefficients $a$ and $b$. We generate multiple initial conditions as a Gaussian random field and multiple forcing functions $F(u)$ using the Fisher-KPP general form with three reaction terms from the literature, as listed in Table \ref{tab:Fisher-KPP}.

\begin{figure}[h!]
\begin{center}
\includegraphics[width=1.0\textwidth]{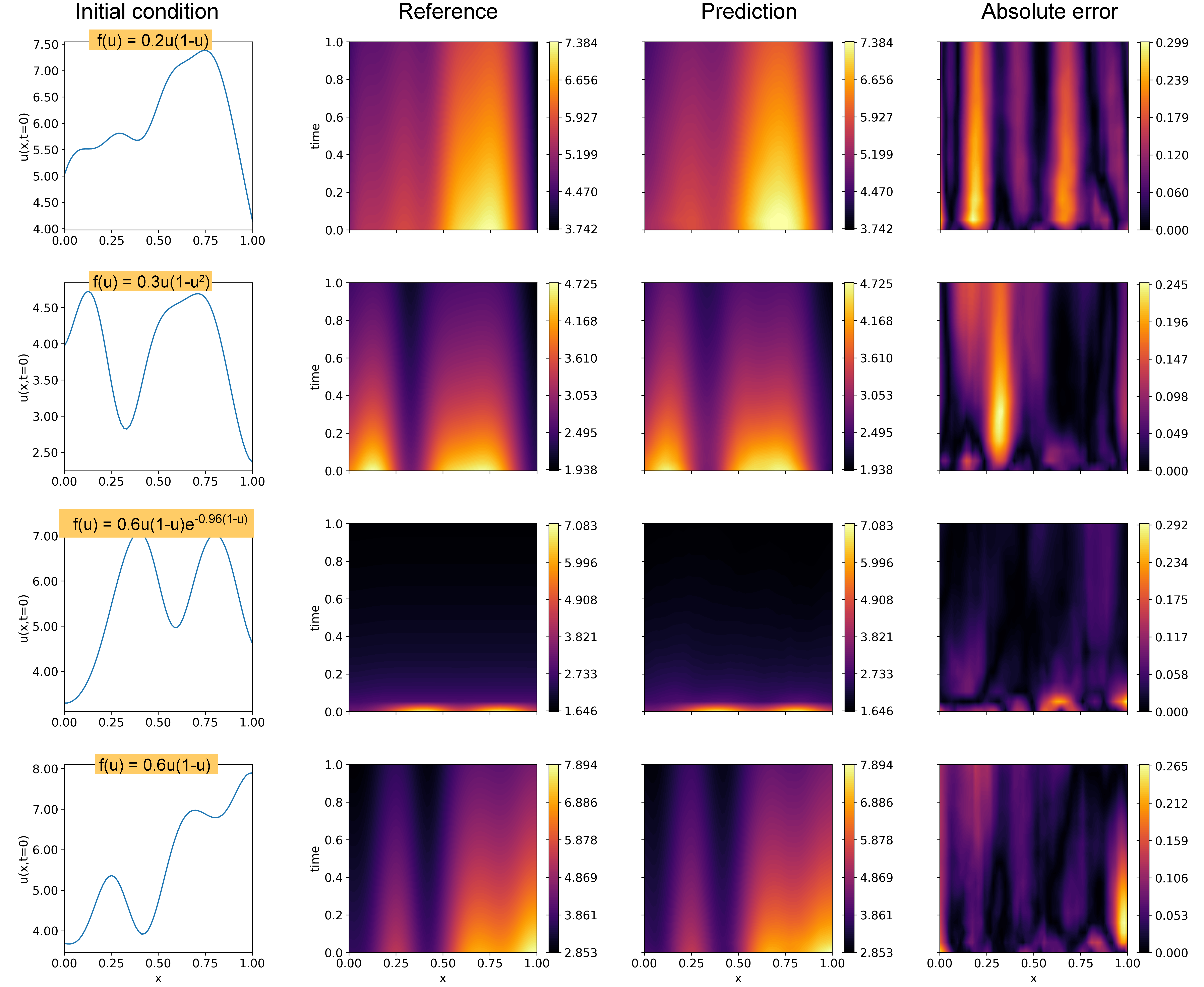}
\caption{Comparison between the reference solution and prediction obtained using the proposed multi-task operator network for representative test cases. The plot shows the operator network predictions against reference solutions for different initial conditions and forcing functions $F(u)$. The results demonstrate good overall accuracy across different initial conditions and forcing functions.}
\label{fig:Fisher_results}
\end{center}
\captionsetup{justification=centering}
\end{figure}

To incorporate the different functional forms of $F(u, a, b)$ as inputs to the network, we express the functions as shown in Equation \ref{eqn:Fisher_poly}. For the Zeldovich form, we use the Taylor expansion and re-write it as:
\begin{align}
    F(u) = au(1-u)e^{-b(1-u)} = (au-au^2)\left( 1 - b(1-u) + \frac{b^2(1-u)^2}{2} + \mathcal{O}(u^3)\right).
\end{align}
The coefficients of $u$, $u^2$, and $u^3$ are represented as $\alpha$, $\beta$ and $\gamma$, respectively in Figure \ref{fig:schematic1}, with the constant term is denoted as $\delta$. To train the MT-DeepONet for the Fisher-KPP equations, the coefficients of the parameterized form of $F(u, a, b)$ are concatenated with the flattened initial conditions, $u(x, t=0)$ and used as inputs to the branch network. We use Adam optimizer with a progressively reducing learning rate and the network is trained for $100$,$000$ epochs using mean-squared error as the loss function. The relative $\mathcal{L}_{2}$ norm for the test samples is approximately $\sim 2.4\%$. Figure ~\ref{fig:Fisher_results} presents four representative test case predictions using the MT-DeepONet. The results indicate that the multi-task operator network can capture large-scale features across the space-time domain with reasonable accuracy, under varying initial conditions and parameters of the Fisher-KPP equation. Additionally, the training time for individual DeepONet for each forcing term with varying initial conditions was $1250$ - $1300$ seconds on an NVIDIA A100 GPU. In contrast, the MT-DeepONet framework was trained in $\approx 1200$ seconds, showing the computational effectiveness of our approach.

\subsection{Darcy Flow in $2$D geometries}
\label{subsec:darcy}

In the second example, we consider the Darcy flow through a bounded domain and aim to train multiple $2$D geometries (bounded domains) simultaneously considering parameterized spatially varying conductivity fields. An earlier attempt, as noted in \cite{he2024geom}, focused on learning similar yet parameterized geometries in a concurrent training session. In contrast, our work demonstrates the learning of distinctly different geometries.

\begin{figure}[h!] 
    \centering
    \begin{subfigure}[t]{0.22\textwidth}
        \centering
        \centerline{\includegraphics[width = 1\textwidth]{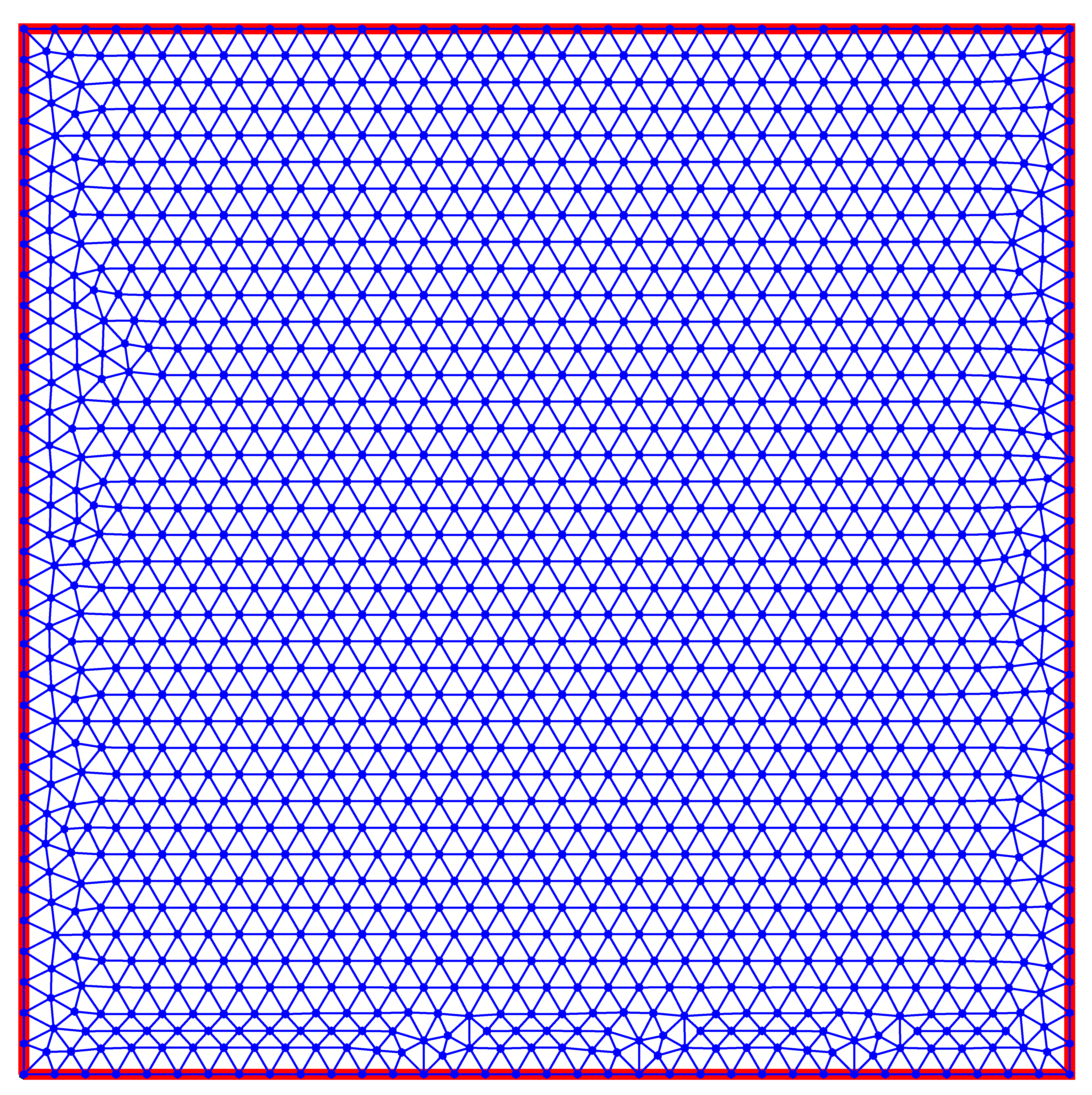}}
        \caption{\textbf{S$1$}: Square}         
    \end{subfigure}
    \hspace*{5pt}
    \begin{subfigure}[t]{0.22\textwidth}
        \centering
        \centerline{\includegraphics[width = 1\textwidth]{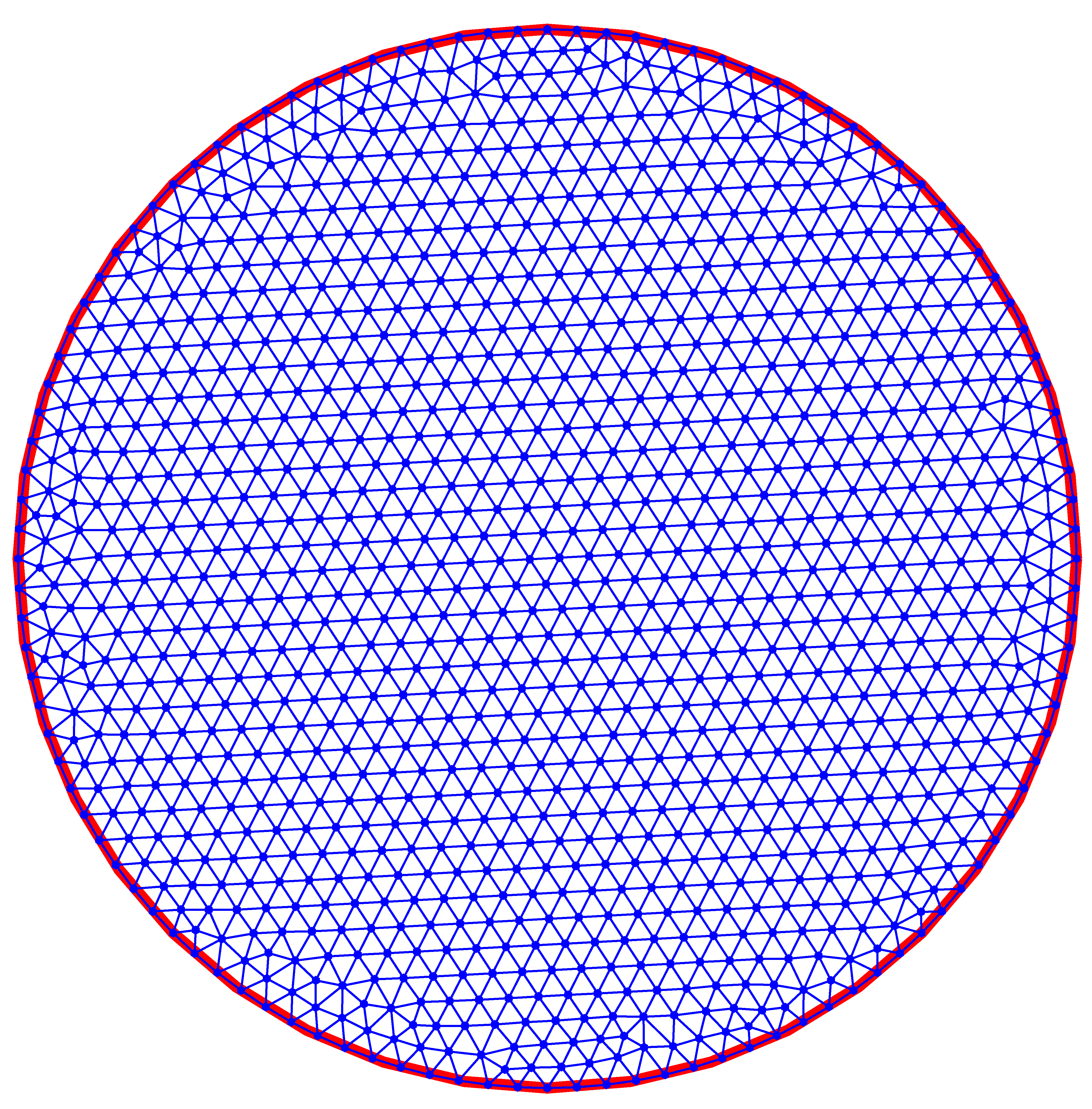}}
        \caption{\textbf{S$2$}: Circle}         
    \end{subfigure}
    \hspace*{5pt}
    \begin{subfigure}[t]{0.22\textwidth}
        \centering
        \centerline{\includegraphics[width = 1\textwidth]{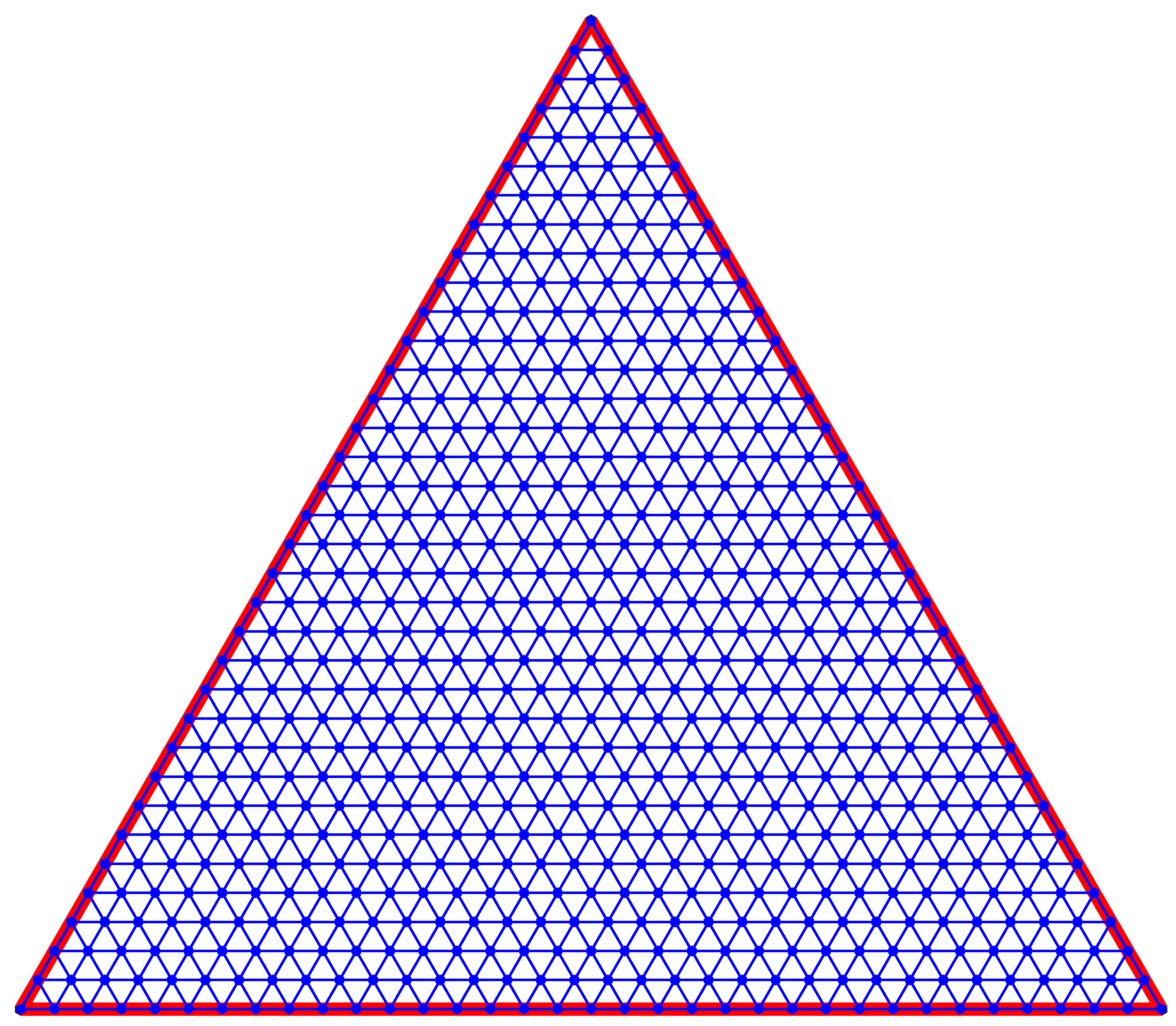}}
        \caption{\textbf{S$3$}: Triangle}         
    \end{subfigure}
    \hspace*{5pt}
    \begin{subfigure}[t]{0.22\textwidth}
        \centering
        \centerline{\includegraphics[width = 1\textwidth]{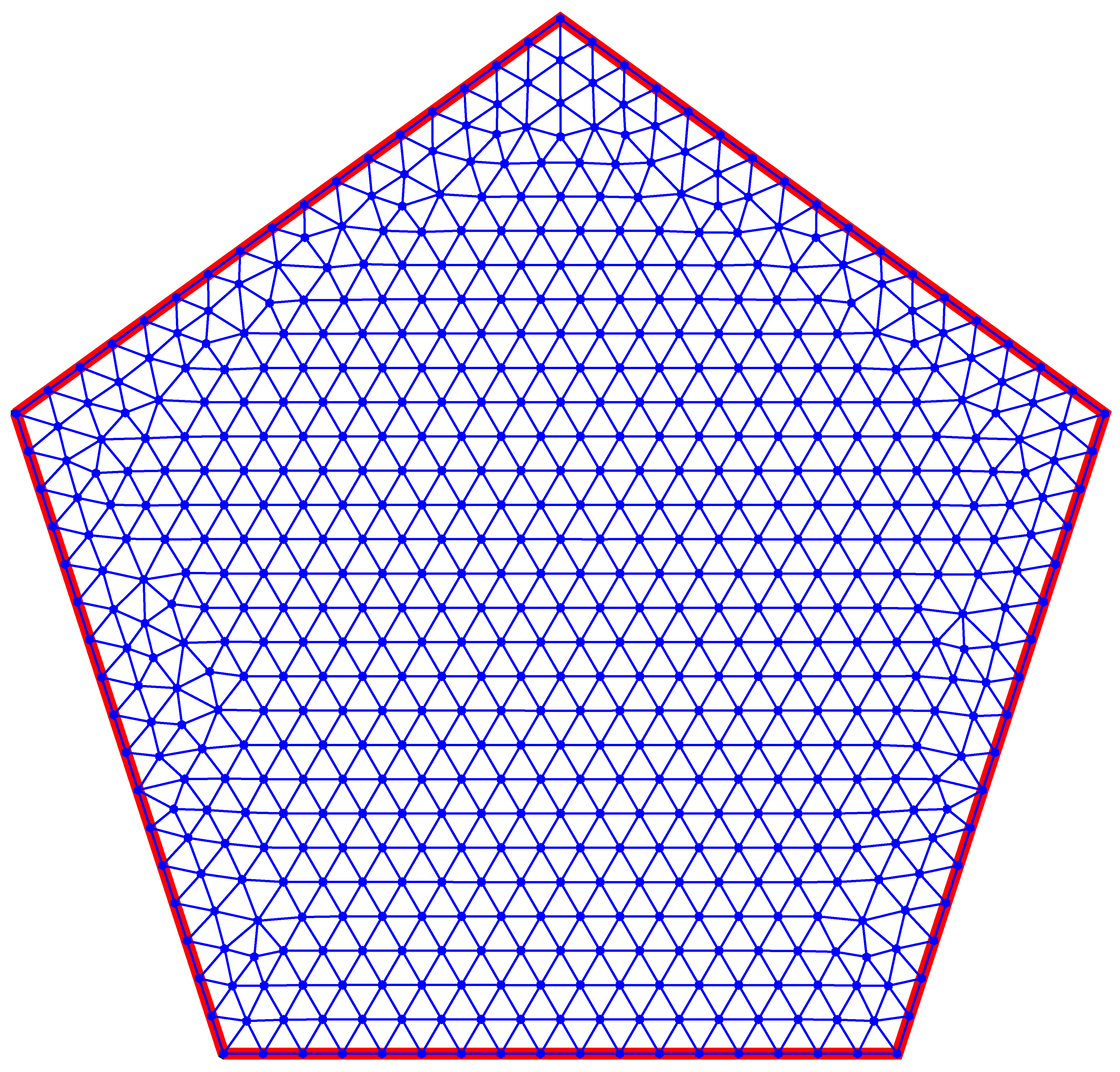}}
        \caption{\textbf{S$4$}: Pentagon}         
    \end{subfigure}
    \\
    \begin{subfigure}[t]{0.22\textwidth}
        \centering
        \centerline{\includegraphics[width = 1\textwidth]{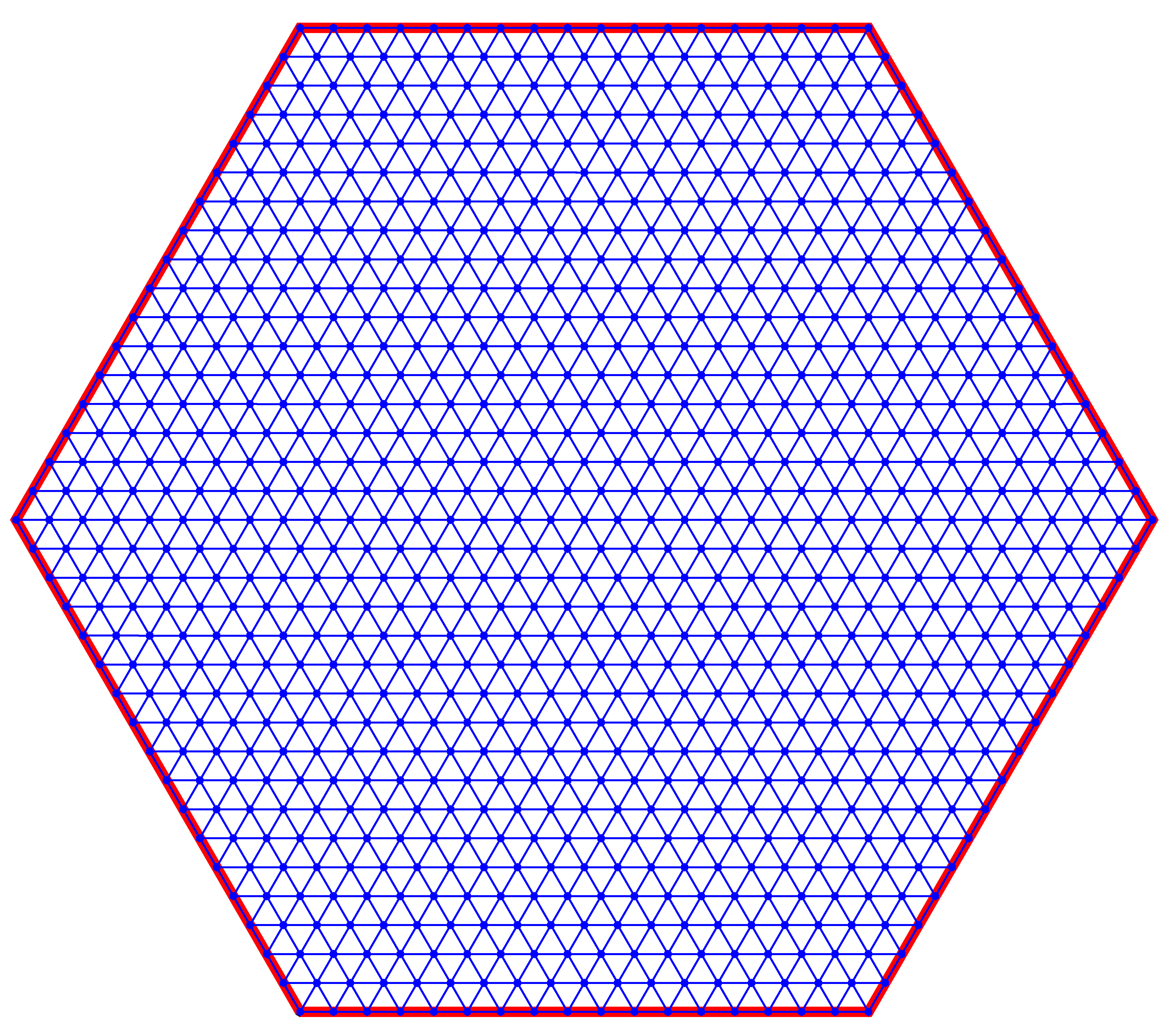}}
        \caption{\textbf{S$5$}: Hexagon}         
    \end{subfigure}
    \hspace*{10pt}
    \begin{subfigure}[t]{0.22\textwidth}
        \centering
        \centerline{\includegraphics[width = 1\textwidth]{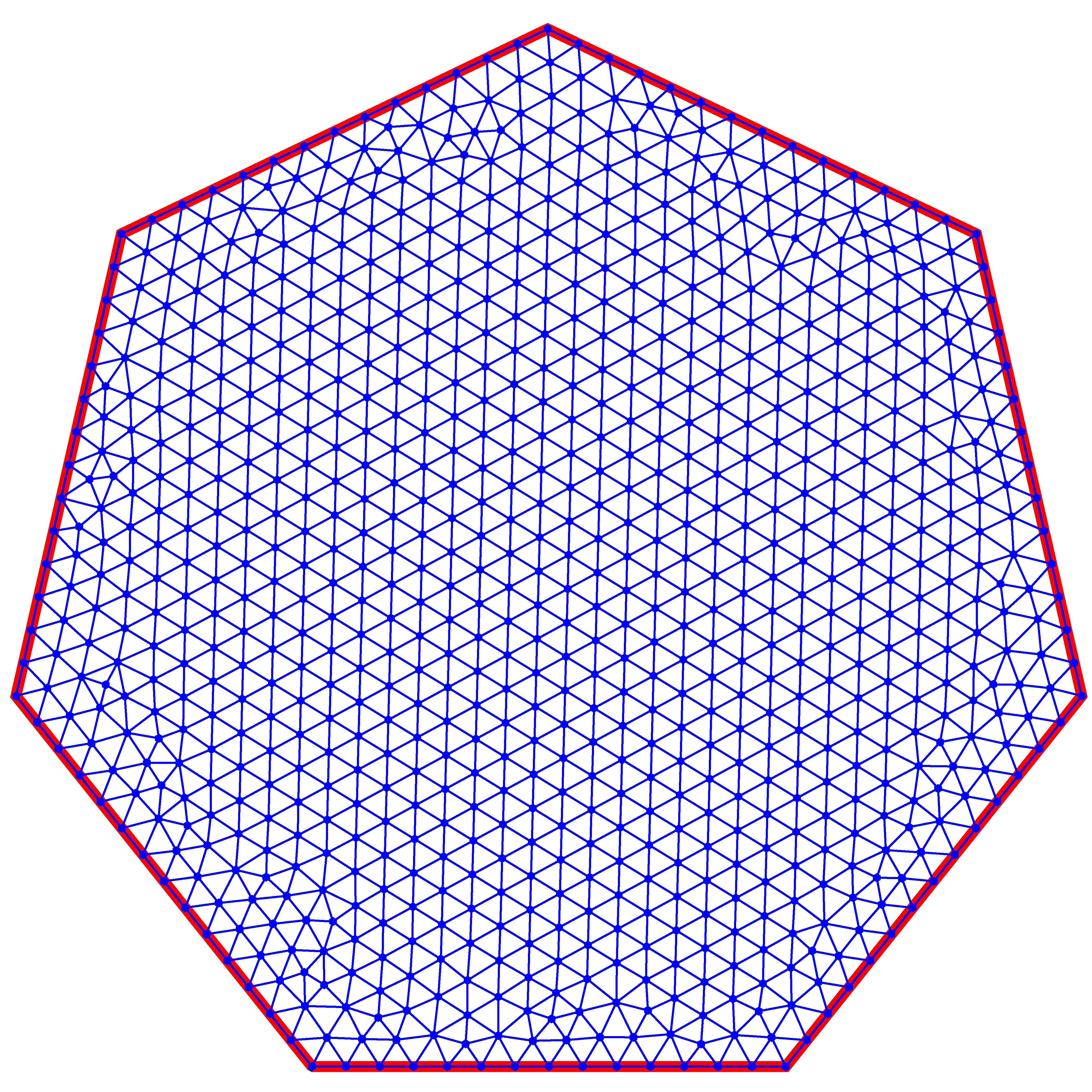}}
        \caption{\textbf{S$6$}: Heptagon}         
    \end{subfigure}
    \hspace*{10pt}
    \begin{subfigure}[t]{0.22\textwidth}
        \centering
        \centerline{\includegraphics[width = 1\textwidth]{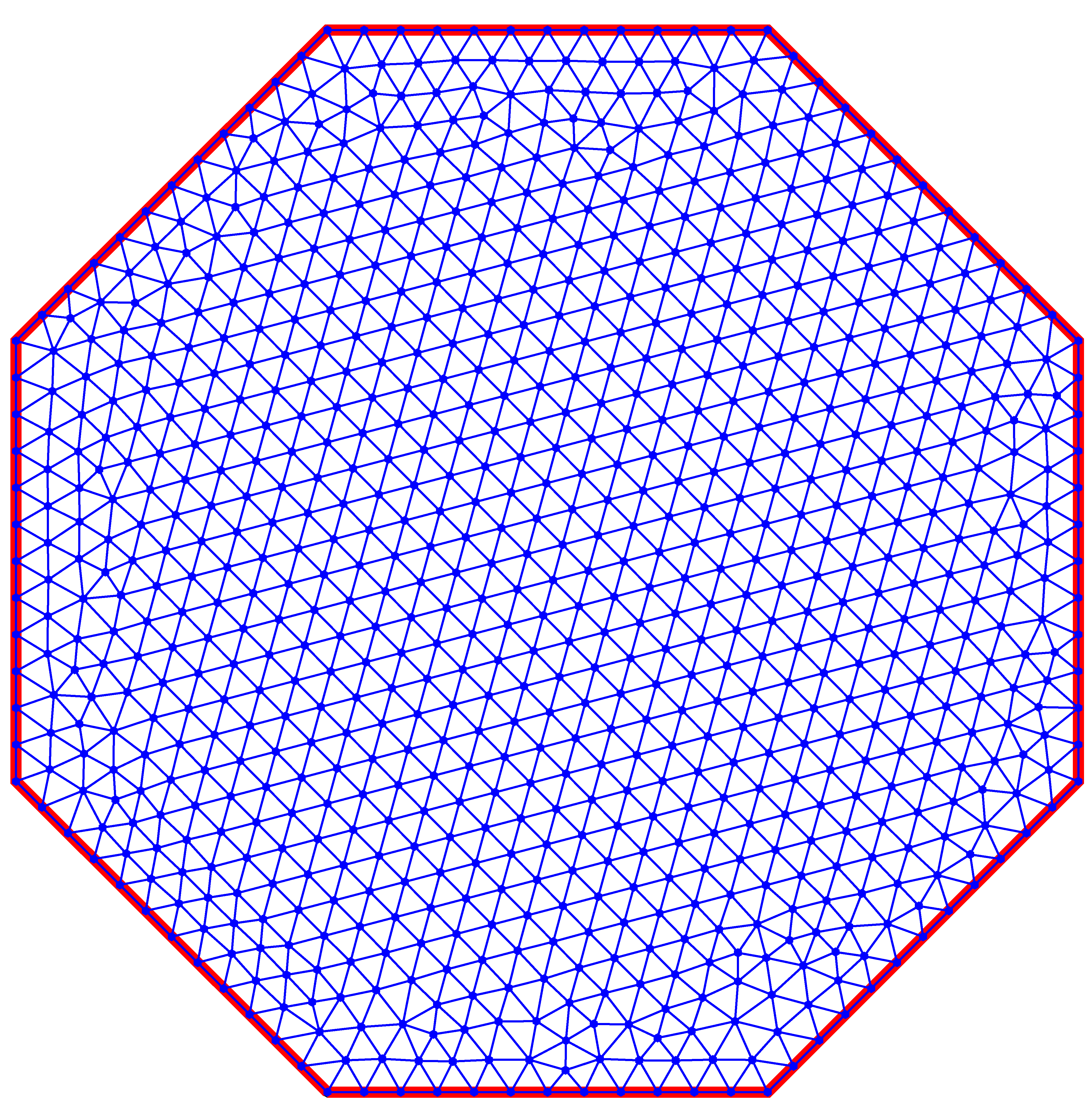}}
        \caption{\textbf{S$7$}: Octagon}         
    \end{subfigure}
    \\
    \begin{subfigure}[t]{0.22\textwidth}
        \centering
        \centerline{\includegraphics[width = 1\textwidth]{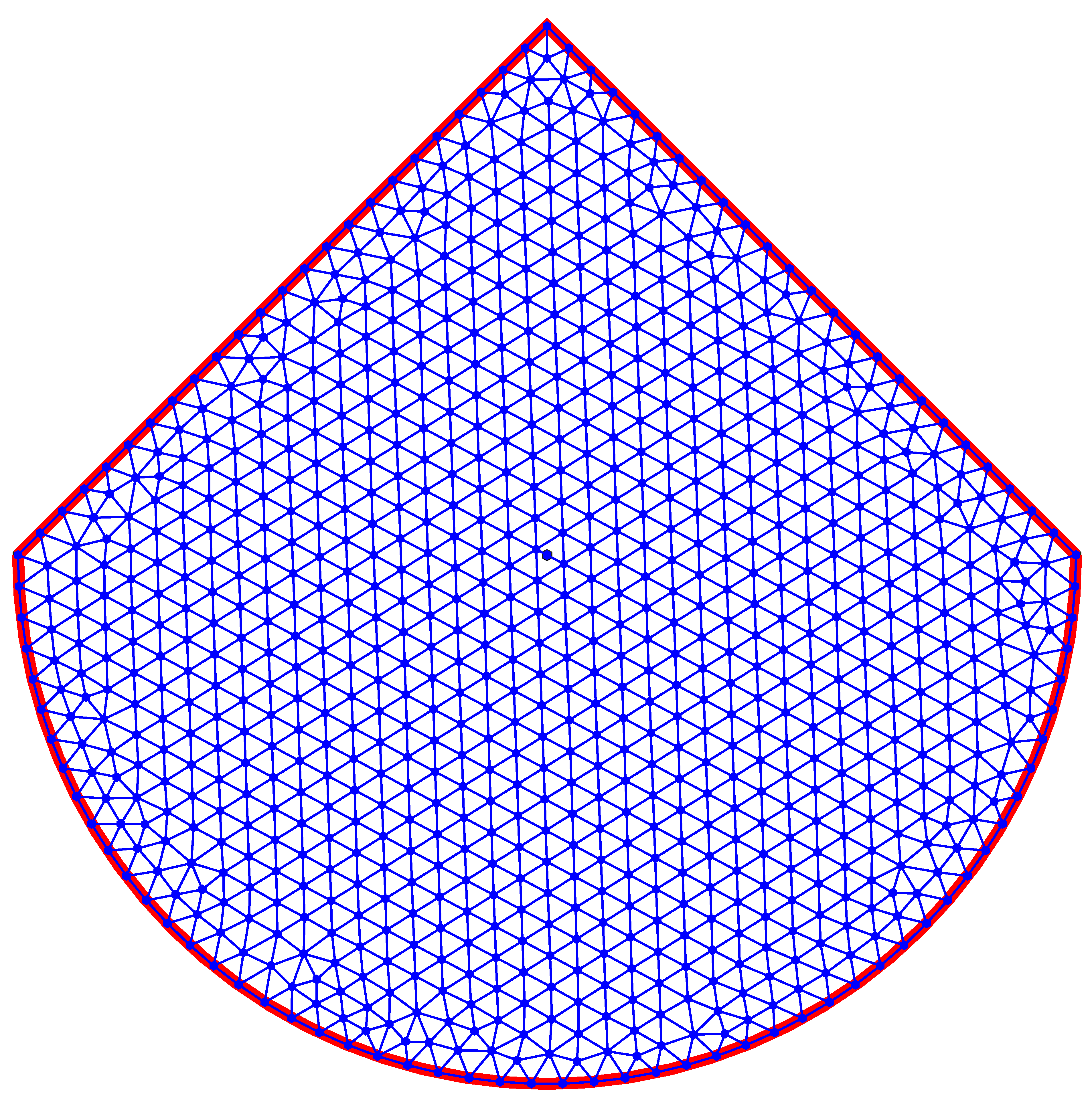}}
        \caption{\textbf{T$1$}: Semi-circle with triangle}         
    \end{subfigure} \hspace*{10pt}
    \begin{subfigure}[t]{0.22\textwidth}
        \centering
        \centerline{\includegraphics[width = 1\textwidth]{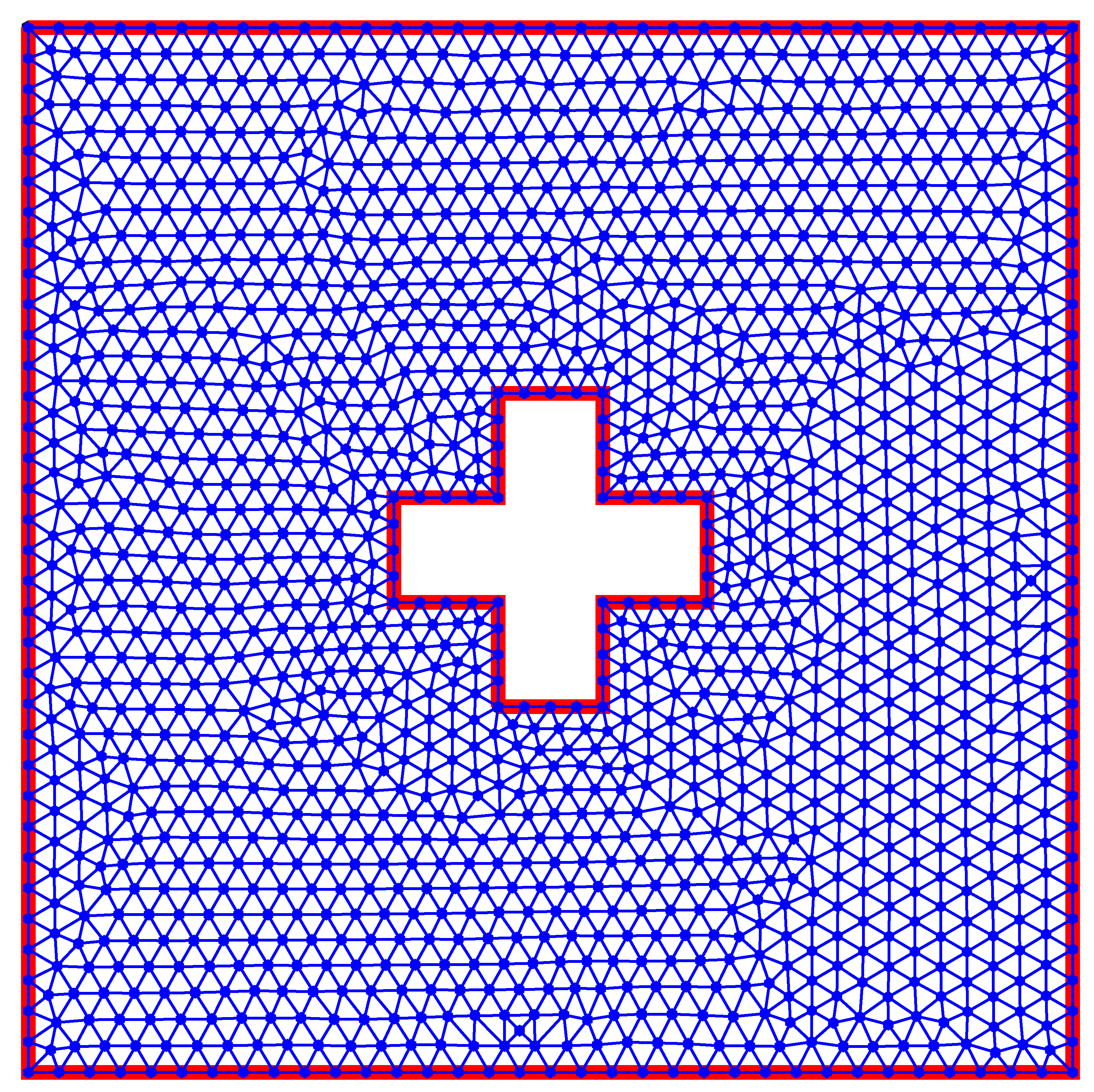}}
        \caption{\textbf{T$2$}: Square with cutout}         
    \end{subfigure}
    \hspace*{10pt}
    \begin{subfigure}[t]{0.22\textwidth}
        \centering
        \centerline{\includegraphics[width = 1\textwidth]{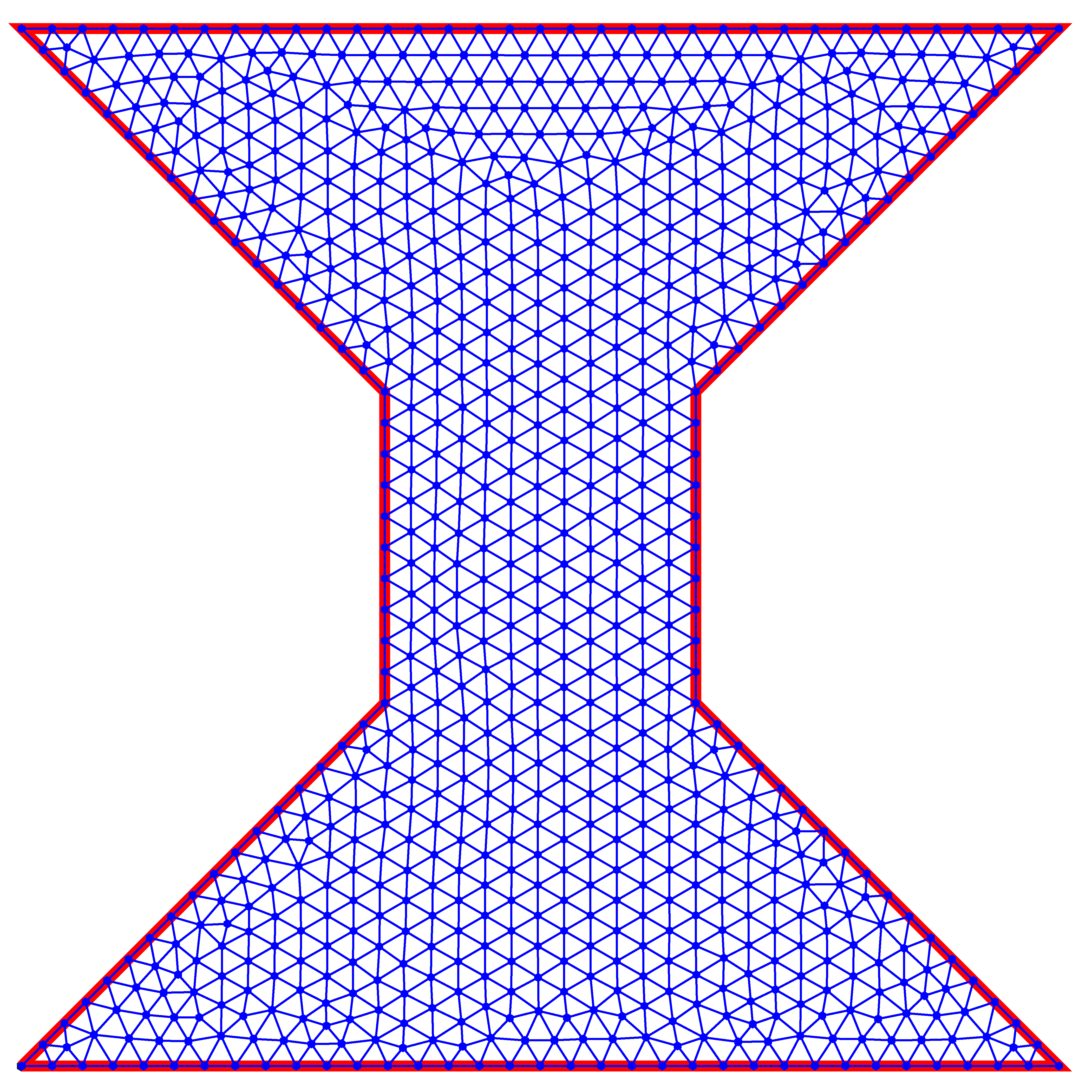}}
        \caption{\textbf{T$3$}: I-section}         
    \end{subfigure}
    \caption{Different geometries considered as tasks for MT-DeepONet for the Darcy flow problem. The source MT-DeepONet is trained using various combinations of the geometries S$1$ - S$7$ while the target domain geometries considered are T$1$ - T$3$. The boundary indicated in red denotes the Dirichlet boundaries where the solution $h(x) = 0$. The first objective of MT-DeepONet is to learn the hydraulic pressure heads across a combination of source geometries given a parametric family of spatially varying conductivity fields. The second objective is to transfer the knowledge of source MT-DeepONet to different geometries in the target domain using the transfer learning approach proposed in \cite{goswami2022transfer_learning} to reduce overall computation time.}
    \label{fig:2D_Darcy_geoms}
\end{figure}

Darcy's law describes fluid flow through a porous medium, relating pressure, velocity, and medium permeability. The pressure in the porous medium is expressed as \cite{goswami2022physics}:
\begin{align}\label{eqn:Darcy}
    & \nabla \cdot (K(\bm{x}) \nabla h(\bm{x})) = g(\bm{x}) \text{ in} \quad \Omega \in \mathbb{R}^2,  \\
    &\text{subject to:  } h(\bm{x}) = 0, \;\; \forall \;\bm{x} \in \partial \Omega,
\end{align}
where $K(\bm{x})$ denotes spatially varying hydraulic conductivity field, $h(\bm{x})$ is the hydraulic head, and $g(\bm{x})$ is the source term. For simplicity, we set $g(\bm{x}) = 1$. The objectives for multi-task operator learning to predict the hydraulic head, $h(\bm{x})$, include ($i$) spatially varying conductivity fields $K(\bm{x})$ drawn from a Gaussian random field, and ($ii$) operation on varying $2$D geometries. In this task, we aim to demonstrate the generalization ability of MT-DeepONet through transfer learning, where the target domains differ geometrically from the source domains. The source and target geometries considered in this example are shown in Figure~\ref{fig:2D_Darcy_geoms}. The source geometries are labeled with \textbf{S}, while the target geometries are labeled with \textbf{T}. The source MT-DeepONet involves training on source domains \textbf{S} concurrently with sufficient labeled data, which is later transferred to related target domains \textbf{T} where only a small amount of training data is available. To capture geometric variation in a single training session, the conductivity field $K(\bm{x})$ is combined with a binary mask $\bm{M_{\text{binary}}}$ and used as input to the branch network, which employs a CNN architecture. The trunk network receives inputs from a uniformly discretized square domain $\Omega \in [0,1] \times [0,1]$, subdivided into a $100 \times 100$ uniform grid. For all geometries, ground truth data is obtained using the MATLAB PDE Toolbox on an irregular mesh and thereafter interpolating the solution onto this $100 \times 100$ regular grid. The trunk network's basis functions (output embeddings) are used to evaluate the solution field at all domain points. The solution field outside the domain boundary is enforced to be zero by multiplying the solution operator with the binary mask. This mask ensures the solution is zero outside the geometry, thus improving convergence. Refer to Algorithm~\ref{alg:masking_algo} showing the implementation details of this workflow.

\bigbreak
\noindent\textbf{Transfer learning across different geometries}\\

\begin{figure}[h!] 
    \begin{center}
    \includegraphics[width=0.95\textwidth]{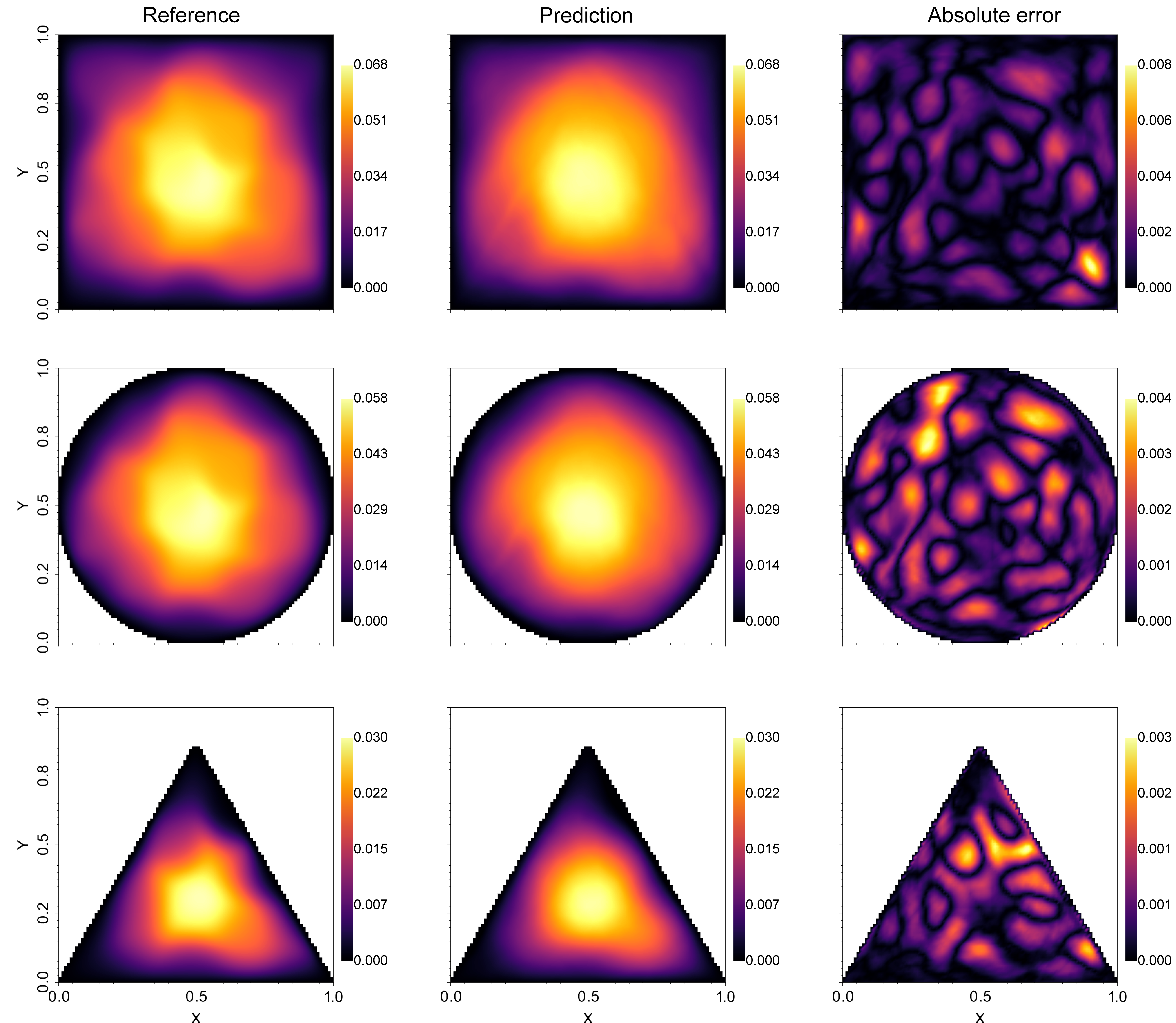}    \caption{Representative prediction results for source model trained on geometries \textbf{S}$1$ + \textbf{S}$2$ + \textbf{S}$3$, concurrently. The MT-DeepONet source model captures the overall features of the solution space across all three geometries reasonably well, making it suitable for use in the transfer learning process.}
    \label{fig:s1+s2+s3}
    \end{center}
\end{figure}

In this section, we investigate how multi-task training in the source model enhances generalization across different geometries in target models. To assess the effectiveness of transfer learning, we train four source models using various geometric combinations: \textbf{S1}, \textbf{S1+S2}, \textbf{S1+S3}, and \textbf{S1+S2+S3}. Each source model is trained on a total of 5,400 samples. For single-geometry models (e.g., \textbf{S1}), all samples are from the same geometry. For multi-geometry models, we evenly distribute samples across the geometries: 2,700 samples from each geometry for \textbf{S1+S2} and \textbf{S1+S3}, and 1,800 samples from each geometry for \textbf{S1+S2+S3}. We train the MT-DeepONet using a piece-wise constant learning rate scheduler with rates [0.001, 0.0005, 0.0001] over 5,000 epochs, employing mini-batching with a batch size of $1,000$ and optimizing with mean squared error. This experimental setup allows us to evaluate how the diversity of geometries in the source model affects the transfer learning process and subsequent performance on target models. Figure \ref{fig:s1+s2+s3} presents three representative cases of predicted solutions when source MT-DeepONet was trained concurrently with geometries \textbf{S}$1$ + \textbf{S}$2$ + \textbf{S}$3$. The binary mask is applied to the output of the operator network to enforce that solutions outside the geometric domain are zero. The results demonstrate that the source MT-DeepONet has accurately learned all three geometries. Table \ref{tab:darcy_results} presents the $\mathcal{L}_{2}$ relative errors for various source model configurations. We observe a trend of increasing error as the number of geometries in the training data grows. This pattern is consistent with our expectations, given the multi-task feature set that the operator network must learn during multi-geometry training. While this approach may lead to a slight reduction in accuracy for individual tasks, it offers a more versatile representation across multiple geometries. This trade-off between task-specific performance and cross-geometry generalization is a key aspect of our multi-task learning strategy.

\begin{figure}[htpb!]
\begin{center}
\includegraphics[width=1\linewidth]{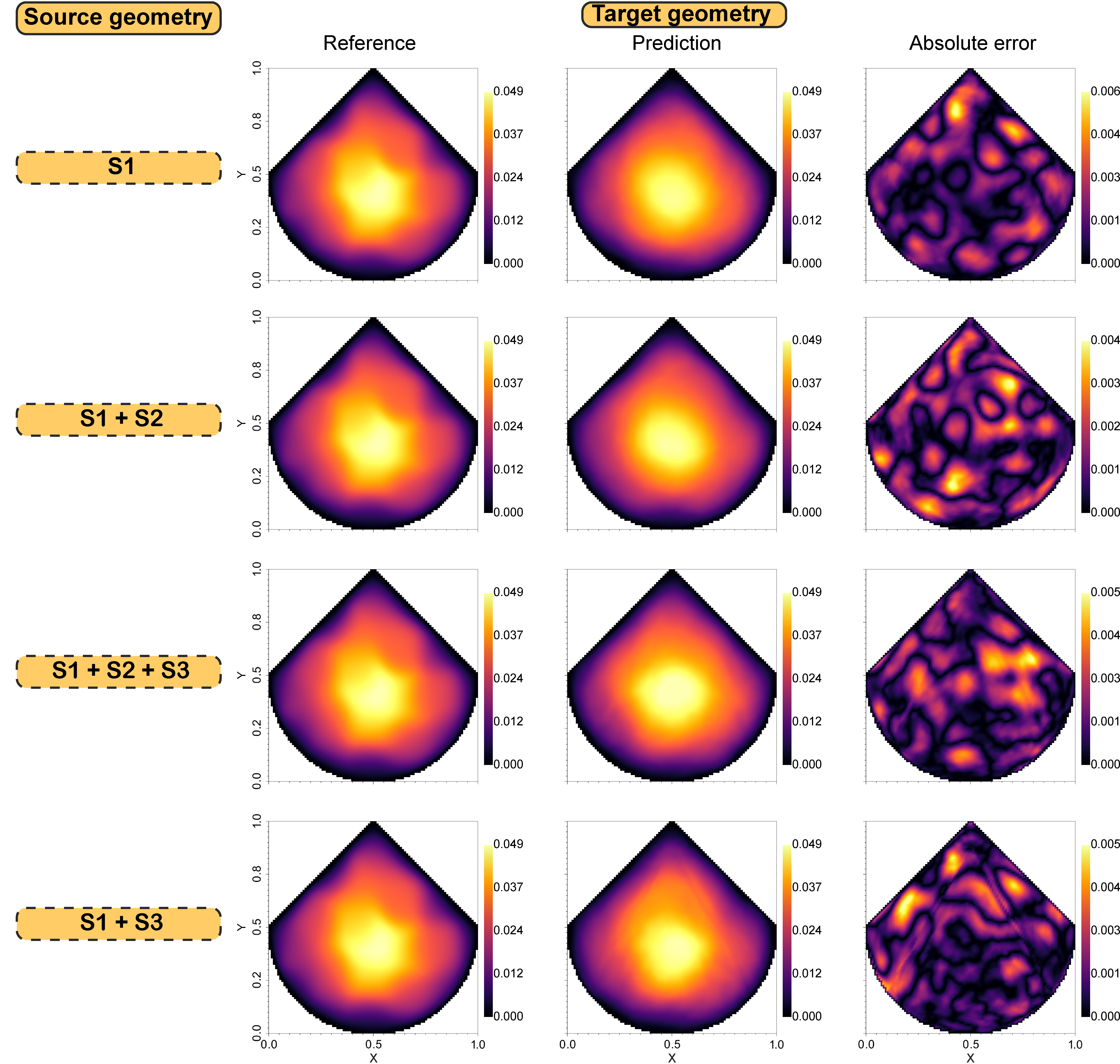}
\caption{Comparative analysis between reference solution and prediction generated by the target model \textbf{T1} trained with 50 samples. Source model \textbf{S1} + \textbf{S2} results in lower error values than source model $S1$ for this geometry. Other source models using \textbf{S3} in training samples lead to higher errors due to negative interference. Note the artifacts that emerge due to negative interference from the triangular source geometry for source model combinations (\textbf{S1} + \textbf{S2} + \textbf{S3}) and (\textbf{S1} + \textbf{S3}). Overall, the target model is capable of capturing the pressure distribution for new geometries relatively well with the transfer learning methodology when appropriate geometries are chosen in the source model.}
\label{fig:Darcy_solution_cone}
\end{center}
\captionsetup{justification=centering}
\end{figure}

\begin{figure}[htpb!]
\begin{center}
\includegraphics[width=1\linewidth]{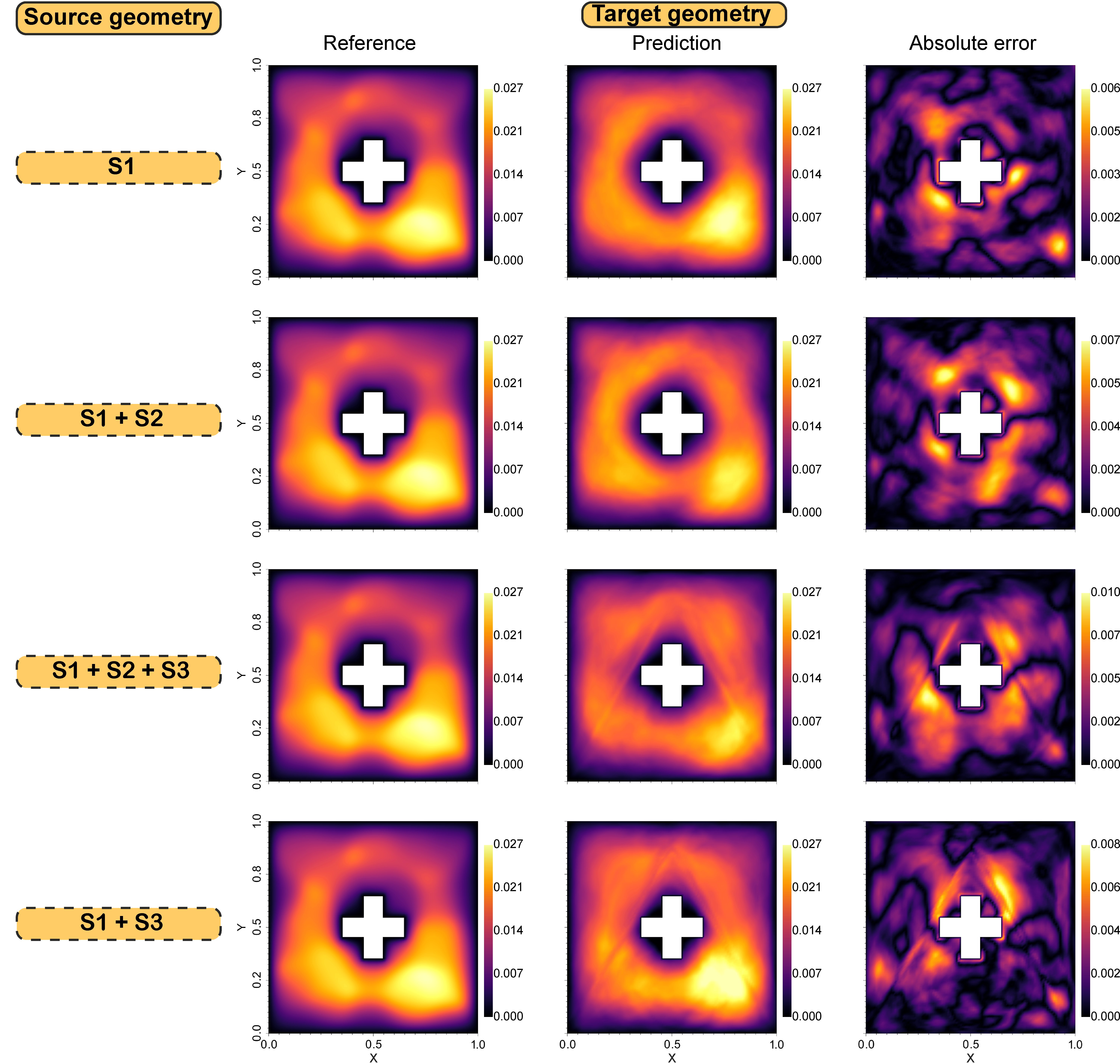}
\caption{Comparative analysis between reference solution and prediction generated by the target model \textbf{T2} fine-tuned with 50 samples. The absolute error plot unveils certain regions with higher error accumulation, especially surrounding the cutout in a square geometric domain (\textbf{T2}). Target geometry \textbf{T2} represents an extrapolated case since none of the source models have an internal Dirichlet boundary with zero pressure head enforced.}
\label{fig:Darcy_solution_cross}
\end{center}
\captionsetup{justification=centering}
\end{figure}

We employ transfer learning to adapt our trained source multi-task operator network for new geometries \textbf{T1} and \textbf{T2}. The target models are considered distinct tasks. The transfer learning process begins by initializing the network with trained parameters from the source model. The layers updated during fine-tuning include the first input CNN layer of the branch network, three MLP layers following the convolution modules in the branch, and the linear output layer of the trunk network. We assess the prediction accuracy of the target model using varying numbers of training samples from the target domain: 50, 100, 200, 500, and 800. This approach significantly reduces computational costs for learning pressure heads on new geometries. Table \ref{tab:darcy_results} presents a summary of errors obtained from the target models for different test cases. Our analysis reveals distinct error patterns for target geometries \textbf{T1} and \textbf{T2}. For \textbf{T1}, the source model combination \textbf{S1} + \textbf{S2} yields a lower $\mathcal{L}_{2}$ error compared to the single-geometry \textbf{S1} source model. This demonstrates that training on multiple geometries can enhance the network's transfer learning capabilities. Conversely, source models with geometrical combinations \textbf{S1} + \textbf{S3} and \textbf{S1} + \textbf{S2} + \textbf{S3} result in higher errors in the target model compared to single-task learning on geometry \textbf{S1}. These findings underscore a crucial insight: naively combining all source tasks does not universally improve prediction performance for the target task. This phenomenon, known as negative transfer, occurs when source tasks unrelated to the target tasks are included. In our case, the results indicate that source geometry \textbf{S3} introduces a negative transfer effect. Negative interference is a known challenge in MTL and we plan to explore mitigation strategies in future research. Conversely, for geometry \textbf{T2}, the source model \textbf{S1} yields similar error values to the combination \textbf{S1} + \textbf{S2}, indicating no significant improvement with the multi-task source model. This result may be attributed to the similarity between geometry $\textbf{T2}$ and $\textbf{S1}$. Figures \ref{fig:Darcy_solution_cone} and \ref{fig:Darcy_solution_cross} compare predictions from the transfer learning model against reference solutions for different source models trained with $50$ samples used to fine-tune the target model. The source model was trained with $5400$ samples where each geometry had equal representation.

\begin{table}[h]
\small
\caption{Relative $\mathcal L_2$ error for TL-DeepONet for different source models vs number of samples used during fine-tuning of the target domains, \textbf{T1} and \textbf{T2}.}
\label{tab:darcy_results}
\resizebox{\columnwidth}{!}{%
\begin{tabular}{|l|c|c|ccccc|}
\hline
\multirow{2}{*}{\textbf{\begin{tabular}[c]{@{}c@{}}Source geometry\\ model\end{tabular}}} &
  \multirow{2}{*}{\textbf{\begin{tabular}[c]{@{}c@{}}Source model \\ $\mathcal{L}_{2}$ rel error\end{tabular}}} &
  \multirow{2}{*}{\textbf{\begin{tabular}[c]{@{}c@{}}Target \\ geometry\end{tabular}}} &
  \multicolumn{5}{c|}{\textbf{\begin{tabular}[c]{@{}c@{}}$\mathcal{L}_{2}$ rel error for\\ target sample size, n\end{tabular}}} \\ \cline{4-8} 
 &
   &
   &
  \multicolumn{1}{c|}{\textbf{n = 50}} &
  \multicolumn{1}{c|}{\textbf{n = 100}} &
  \multicolumn{1}{c|}{\textbf{n = 200}} &
  \multicolumn{1}{c|}{\textbf{n = 500}} &
  \textbf{n = 800} \\ \hline
\textbf{S1} &
  0.029 &
  \textbf{T1} &
  \multicolumn{1}{c|}{0.082} &
  \multicolumn{1}{c|}{0.063} &
  \multicolumn{1}{c|}{0.055} &
  \multicolumn{1}{c|}{0.047} &
  0.045 \\ \hline
\textbf{S1+S2} &
  0.039 &
  \textbf{T1} &
  \multicolumn{1}{c|}{0.065} &
  \multicolumn{1}{c|}{0.054} &
  \multicolumn{1}{c|}{0.048} &
  \multicolumn{1}{c|}{0.047} &
  0.044 \\ \hline
\textbf{S1+S3} &
  0.061 &
  \textbf{T1} &
  \multicolumn{1}{c|}{0.088} &
  \multicolumn{1}{c|}{0.080} &
  \multicolumn{1}{c|}{0.070} &
  \multicolumn{1}{c|}{0.064} &
  0.061 \\ \hline  
\textbf{S1+S2+S3} &
  0.054 &
  \textbf{T1} &
  \multicolumn{1}{c|}{0.083} &
  \multicolumn{1}{c|}{0.068} &
  \multicolumn{1}{c|}{0.060} &
  \multicolumn{1}{c|}{0.055} &
  0.054 \\ \hline
  \textbf{S1} &
  0.029 &
  \textbf{T2} &
  \multicolumn{1}{c|}{0.158} &
  \multicolumn{1}{c|}{0.132} &
  \multicolumn{1}{c|}{0.106} &
  \multicolumn{1}{c|}{0.083} &
  0.077 \\ \hline
\textbf{S1+S2} &
  0.039 &
  \textbf{T2} &
  \multicolumn{1}{c|}{0.163} &
  \multicolumn{1}{c|}{0.137} &
  \multicolumn{1}{c|}{0.105} &
  \multicolumn{1}{c|}{0.085} &
  0.080 \\ \hline
\textbf{S1+S3} &
  0.061 &
  \textbf{T2} &
  \multicolumn{1}{c|}{0.152} &
  \multicolumn{1}{c|}{0.134} &
  \multicolumn{1}{c|}{0.117} &
  \multicolumn{1}{c|}{0.102} &
  0.098 \\ \hline  
\textbf{S1+S2+S3} &
  0.054 &
  \textbf{T2} &
  \multicolumn{1}{c|}{0.158} &
  \multicolumn{1}{c|}{0.138} &
  \multicolumn{1}{c|}{0.119} &
  \multicolumn{1}{c|}{0.098} &
  0.094 \\ \hline
\end{tabular}}
\end{table}

\begin{figure}[b!]
\begin{center}
\includegraphics[width=0.95\linewidth]{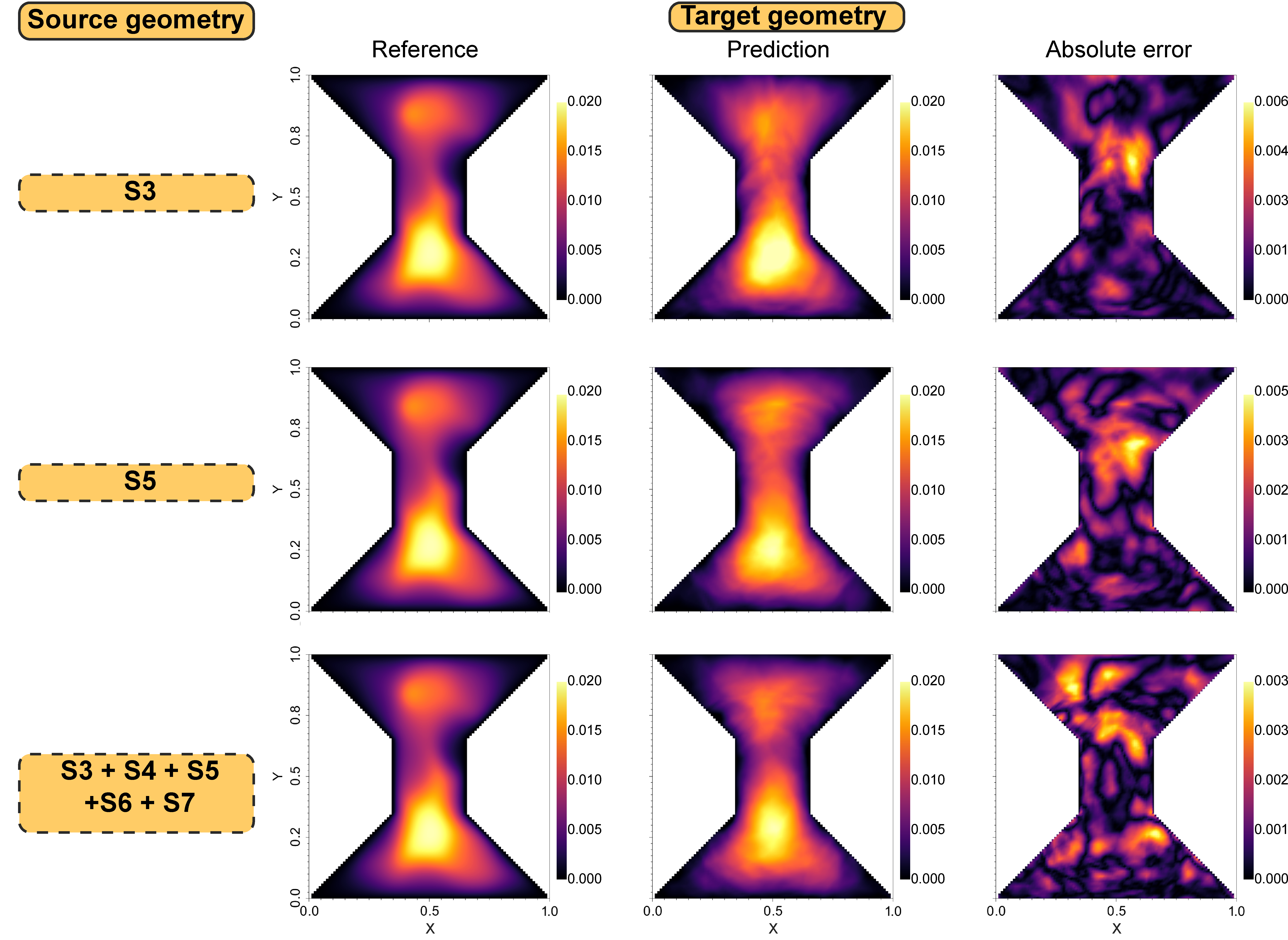}
\caption{A comparison between the reference solution and the prediction from our multi-task operator network for the transfer learning model, trained with $50$ samples on geometry combinations of \textbf{S3-S7}. The I-section geometry used as the target is significantly different from the geometries used in the source model. Utilizing a source model trained on multiple geometries for transfer learning results in a marginally lower prediction error which may not be significant.}
\label{fig:Darcy_sol_add_geoms}
\end{center}
\captionsetup{justification=centering}
\end{figure}

\bigbreak
\noindent\textbf{Transfer learning in the context of disparate geometric domain, \textbf{T3}}\\
Our next objective is to investigate whether incorporating multiple geometries in the source model helps reduce the prediction error when transferring to target geometries with different external boundaries. To further explore the efficacy of multi-geometry training in transfer learning, we introduce an I-section geometry (\textbf{T3}) as our target for transfer learning. Our hypothesis posits that a source model trained on multiple geometries simultaneously can more effectively transfer knowledge to a new target domain. To validate this hypothesis, we compare the prediction errors of transfer learning models derived from single-geometry source models against the one trained on multiple geometries. Table \ref{tab:darcy_additional_results} outlines the source geometry combinations employed in this study: \textbf{S3}, \textbf{S5}, \textbf{S7}, and the multi-geometry combination \textbf{S3-S7}. For consistency, each source model was trained on a total of 5,400 samples. The results, also presented in Table \ref{tab:darcy_additional_results}, reveal that the lowest prediction error for target geometry \textbf{T3} is achieved when the source model is trained on multiple geometries simultaneously. Notably, the improvement in prediction accuracy is more pronounced when the number of target domain training samples is low, with the performance gap narrowing as the number of target domain samples increases. Figure \ref{fig:Darcy_sol_add_geoms} provides a visual comparison of MT-DeepONet predictions against reference solutions for target geometry \textbf{T3}, using three distinct source geometry configurations. While all models demonstrate the ability to capture high-level features, a marked improvement in model prediction is observed when the source model is trained with multiple geometries. This visual evidence corroborates our quantitative findings and underscores the potential benefits of multi-geometry training in enhancing the transferability and generalization capabilities of our model.

\bigbreak
\noindent\textbf{Transfer learning with fine tuning of additional trunk layer}\\
In previous discussions, we presented findings on transfer learning across different $2$D domains, focusing on fine-tuning the last trunk layer along with the first CNN module and the three MLP layers within the branch network. We also evaluated the impact of increasing the number of trainable parameters during the transfer learning phase on the model's accuracy. Specifically, we trained the last two layers of the trunk network (the output layer and the last non-linear hidden layer). We chose to use an additional trunk layer for fine-tuning based on experiments where we observed better accuracy with an additional trunk layer compared to a CNN layer in a branch network. Empirical evidence from our experiments suggests that retraining the last hidden layer has a marginal impact on computational speed. Table \ref{tab:darcy_additional_results} provides an overview of the errors observed across various source-target combinations. Compared to fine-tuning with a single trunk layer, transfer learning with two trunk layers shows a marginal improvement in prediction accuracy across all combinations, particularly for cases with larger sample sizes during transfer learning. This indicates that using additional trainable parameters during transfer learning does not improve the results significantly.

\begin{table}[]
\small
\caption{Error values for different sample sizes used to fine-tune the target model under the transfer learning scheme for the target domain, \textbf{T3}. The table presents the results for two transfer learning scenarios, each differing in the layers selected for fine-tuning.}
\label{tab:darcy_additional_results}
\resizebox{\columnwidth}{!}{%
\begin{tabular}{|lccccccc|}
\hline
\multicolumn{1}{|c|}{} &
  \multicolumn{1}{c|}{} &
  \multicolumn{1}{c|}{} &
  \multicolumn{5}{c|}{\textbf{\begin{tabular}[c]{@{}c@{}}$\mathcal{L}_{2}$ rel error for\\ target sample size, n\end{tabular}}} \\ \cline{4-8} 
\multicolumn{1}{|c|}{\multirow{-2}{*}{\textbf{\begin{tabular}[c]{@{}c@{}}Source geometry\\ model\end{tabular}}}} &
  \multicolumn{1}{c|}{\multirow{-2}{*}{\textbf{\begin{tabular}[c]{@{}c@{}}Source model \\ $\mathcal{L}_{2}$ rel error\end{tabular}}}} &
  \multicolumn{1}{c|}{\multirow{-2}{*}{\textbf{\begin{tabular}[c]{@{}c@{}}Target \\ geometry\end{tabular}}}} &
  \multicolumn{1}{c|}{\textbf{n = 50}} &
  \multicolumn{1}{c|}{\textbf{n = 100}} &
  \multicolumn{1}{c|}{\textbf{n = 200}} &
  \multicolumn{1}{c|}{\textbf{n = 500}} &
  \textbf{n = 800} \\ \hline
\multicolumn{8}{|c|}{\cellcolor[HTML]{FFC702}\textit{\textbf{Results with fine-tuning 3 branch and 1 trunk layer}}} \\ \hline
\multicolumn{1}{|l|}{S3} &
  \multicolumn{1}{c|}{0.045} &
  \multicolumn{1}{c|}{T3} &
  \multicolumn{1}{c|}{0.177} &
  \multicolumn{1}{c|}{0.163} &
  \multicolumn{1}{c|}{0.142} &
  \multicolumn{1}{c|}{0.126} &
  0.120 \\ \hline
\multicolumn{1}{|l|}{S4} &
  \multicolumn{1}{c|}{0.042} &
  \multicolumn{1}{c|}{T3} &
  \multicolumn{1}{c|}{0.176} &
  \multicolumn{1}{c|}{0.159} &
  \multicolumn{1}{c|}{0.141} &
  \multicolumn{1}{c|}{0.120} &
  0.115 \\ \hline
\multicolumn{1}{|l|}{S5} &
  \multicolumn{1}{c|}{0.043} &
  \multicolumn{1}{c|}{T3} &
  \multicolumn{1}{c|}{0.172} &
  \multicolumn{1}{c|}{0.161} &
  \multicolumn{1}{c|}{0.140} &
  \multicolumn{1}{c|}{0.114} &
  0.107 \\ \hline
\multicolumn{1}{|l|}{S6} &
  \multicolumn{1}{c|}{0.042} &
  \multicolumn{1}{c|}{T3} &
  \multicolumn{1}{c|}{0.187} &
  \multicolumn{1}{c|}{0.164} &
  \multicolumn{1}{c|}{0.153} &
  \multicolumn{1}{c|}{0.129} &
  0.116 \\ \hline
\multicolumn{1}{|l|}{S7} &
  \multicolumn{1}{c|}{0.042} &
  \multicolumn{1}{c|}{T3} &
  \multicolumn{1}{c|}{0.183} &
  \multicolumn{1}{c|}{0.166} &
  \multicolumn{1}{c|}{0.148} &
  \multicolumn{1}{c|}{0.128} &
  0.117 \\ \hline
\multicolumn{1}{|l|}{S3+S4+S5+S6+S7} &
  \multicolumn{1}{c|}{0.063} &
  \multicolumn{1}{c|}{T3} &
  \multicolumn{1}{c|}{0.167} &
  \multicolumn{1}{c|}{0.154} &
  \multicolumn{1}{c|}{0.138} &
  \multicolumn{1}{c|}{0.125} &
  0.116 \\ \hline
\multicolumn{8}{|c|}{\cellcolor[HTML]{FFCB2F}\textit{\textbf{Results with fine tuning 3 branch and 2 trunk layers}}} \\ \hline
\multicolumn{1}{|l|}{S3} &
  \multicolumn{1}{c|}{0.045} &
  \multicolumn{1}{c|}{T3} &
  \multicolumn{1}{c|}{0.158} &
  \multicolumn{1}{c|}{0.143} &
  \multicolumn{1}{c|}{0.124} &
  \multicolumn{1}{c|}{0.102} &
  0.090 \\ \hline
\multicolumn{1}{|l|}{S4} &
  \multicolumn{1}{c|}{0.042} &
  \multicolumn{1}{c|}{T3} &
  \multicolumn{1}{c|}{0.165} &
  \multicolumn{1}{c|}{0.143} &
  \multicolumn{1}{c|}{0.118} &
  \multicolumn{1}{c|}{0.097} &
  0.088 \\ \hline
\multicolumn{1}{|l|}{S5} &
  \multicolumn{1}{c|}{0.043} &
  \multicolumn{1}{c|}{T3} &
  \multicolumn{1}{c|}{0.181} &
  \multicolumn{1}{c|}{0.148} &
  \multicolumn{1}{c|}{0.125} &
  \multicolumn{1}{c|}{0.100} &
  0.091 \\ \hline
\multicolumn{1}{|l|}{S6} &
  \multicolumn{1}{c|}{0.042} &
  \multicolumn{1}{c|}{T3} &
  \multicolumn{1}{c|}{0.189} &
  \multicolumn{1}{c|}{0.157} &
  \multicolumn{1}{c|}{0.133} &
  \multicolumn{1}{c|}{0.109} &
  0.098 \\ \hline
\multicolumn{1}{|l|}{S7} &
  \multicolumn{1}{c|}{0.042} &
  \multicolumn{1}{c|}{T3} &
  \multicolumn{1}{c|}{0.181} &
  \multicolumn{1}{c|}{0.153} &
  \multicolumn{1}{c|}{0.141} &
  \multicolumn{1}{c|}{0.111} &
  0.101 \\ \hline
\multicolumn{1}{|l|}{S3+S4+S5+S6+S7} &
  \multicolumn{1}{c|}{0.063} &
  \multicolumn{1}{c|}{T3} &
  \multicolumn{1}{c|}{0.161} &
  \multicolumn{1}{c|}{0.142} &
  \multicolumn{1}{c|}{0.128} &
  \multicolumn{1}{c|}{0.104} &
  0.096 \\ \hline
\end{tabular}%
}
\end{table}

\subsection{Heat transfer through multiple $3$D geometries}
\label{subsec:example4}
\begin{figure}[h!] 
     \centering
     \begin{subfigure}[b]{0.45\textwidth}
         \centering
         \includegraphics[width=0.7\linewidth]{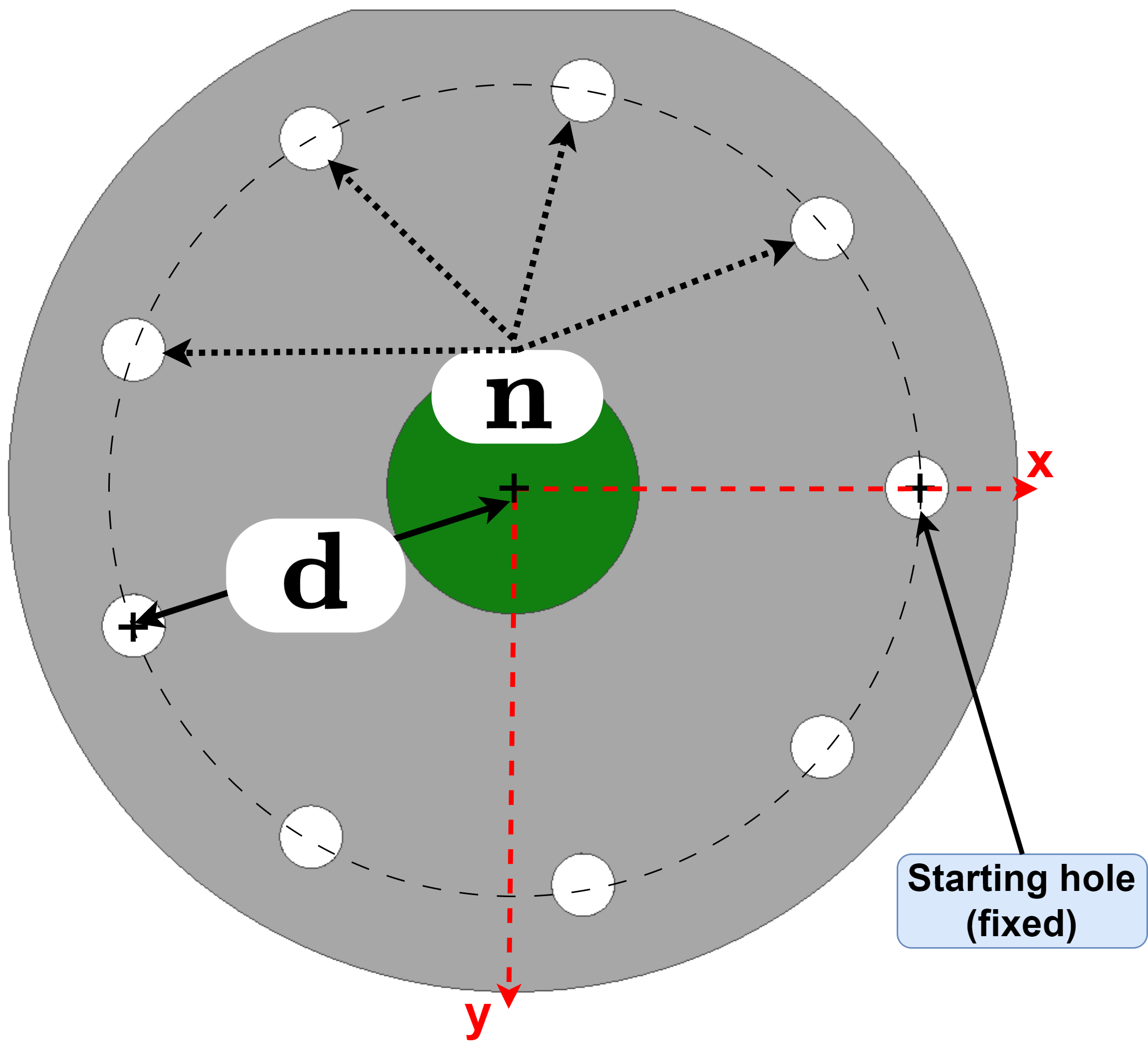}
         \caption{}
         \label{fig:3D_plate_params}
     \end{subfigure}
     \begin{subfigure}[b]{0.5\textwidth}
         \centering
         \includegraphics[width=1\linewidth]{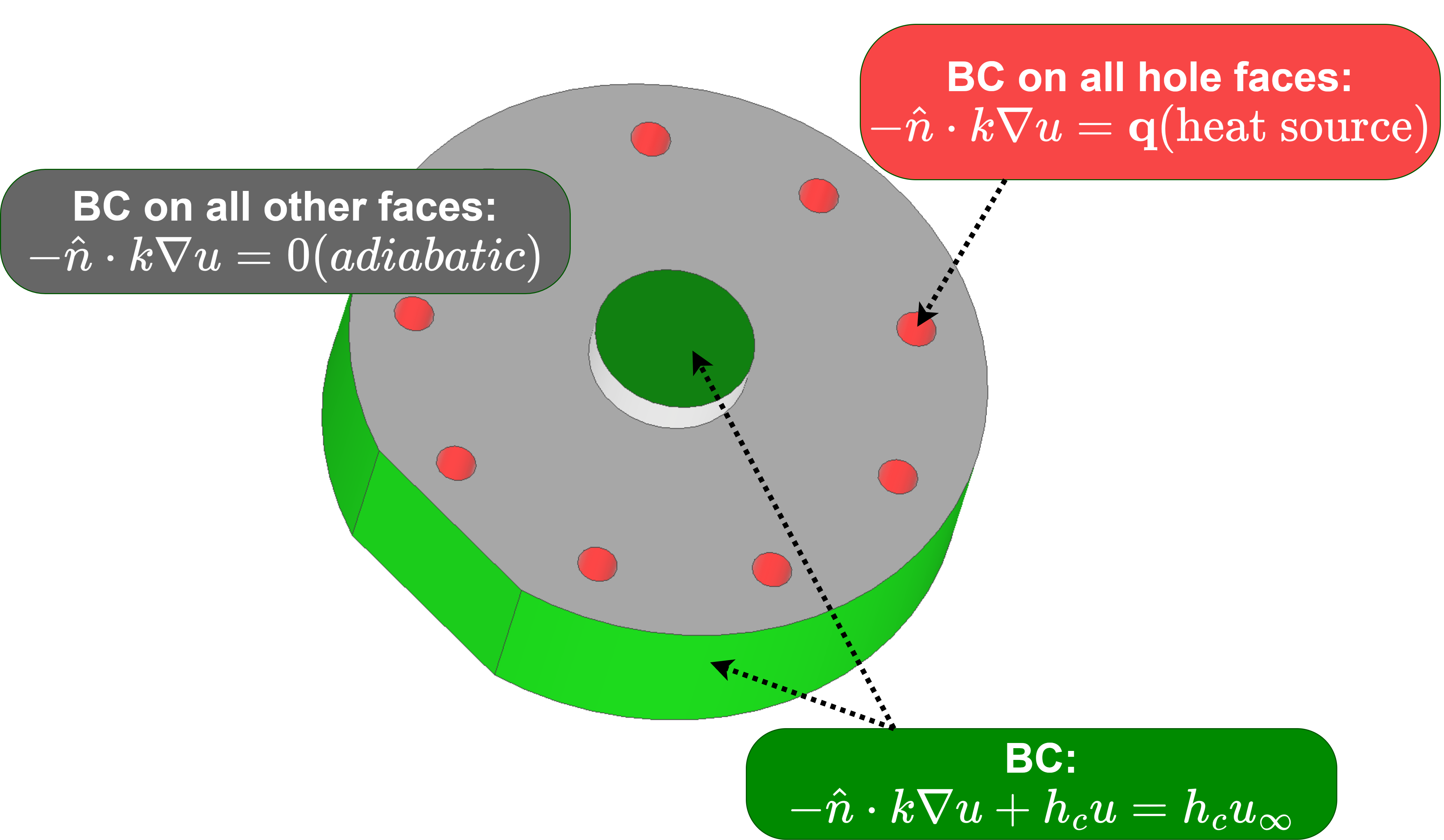}
         \caption{}
         \label{fig:3D_plate_bcs}
     \end{subfigure}
     \caption{(a) View of a $3$D plate showing the parametric design features, where $\mathbf{n}$ represents the number of holes and $\mathbf{d}$ denotes the location of the holes relative to the central axis of the protrusion face. Multiple geometries are generated by varying $\mathbf{n}$ and $\mathbf{d}$ within their design ranges. (b) depicts the boundary conditions used for solving the heat transfer equation. The top protrusion and the side wall have a convective heat flux boundary condition, while the holes act as individual heat sources. The solution field consists of the temperature distribution across the $3$D plate domain.}
     \label{fig:3D_plate}
\end{figure}

\begin{figure}[h!] 
     \centering
     \begin{subfigure}[b]{0.9\textwidth}
         \centering
         \centerline{\includegraphics[width = 1\textwidth]{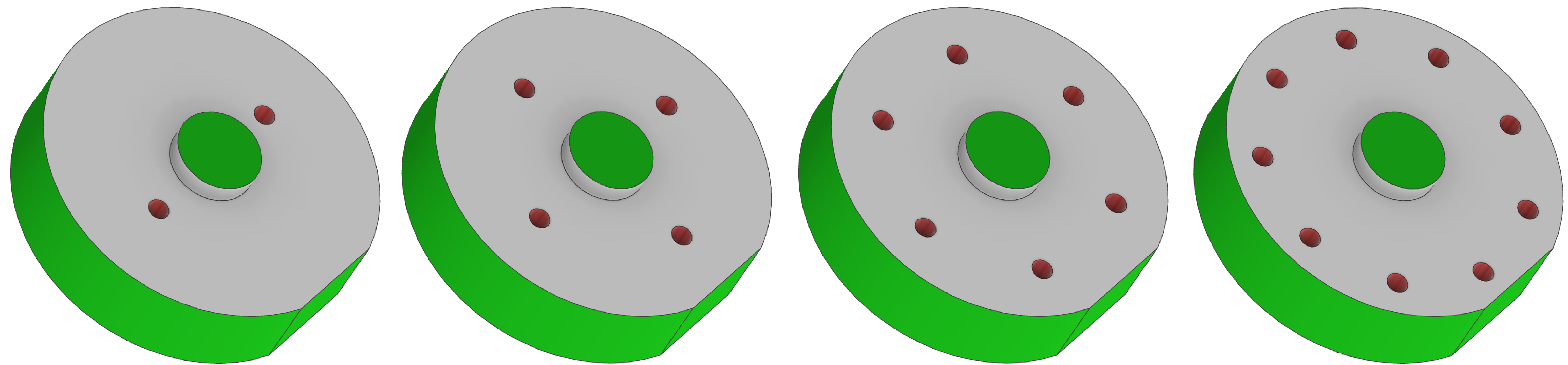}}
         \caption{}         
     \end{subfigure} \\
     \vspace*{5pt}
      \begin{subfigure}[b]{0.9\textwidth}
         \centering
         \centerline{\includegraphics[width = 1\textwidth]{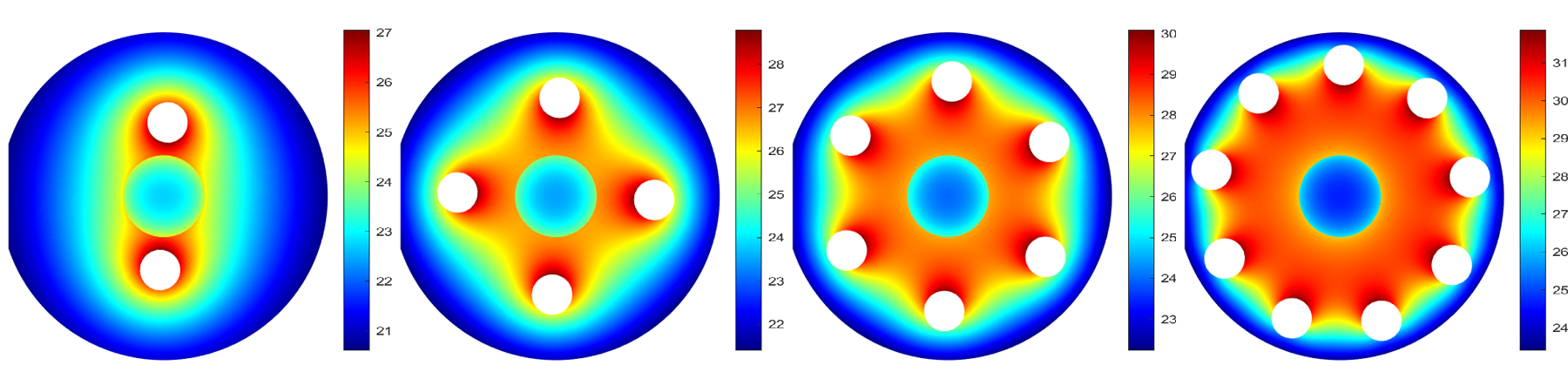}}
         \caption{}         
     \end{subfigure}
       \caption{(a)Representative $3$D plate geometry configurations generated by changing the two design parameters: $\boldsymbol{n}$ and $\boldsymbol{d}$. (b) Variations in temperature fields for different geometric configurations and number of heating sources. The multi-task objective is to learn the mapping from geometric parameters to temperature fields for different geometric configurations.}
       \label{fig:3D_plate_eg}
\end{figure}

The steady-state heat transfer equation for an isotropic medium is defined as:
\begin{align}\label{eqn:heat_transfer}
    -\nabla \cdot (K \nabla \boldsymbol{u}) = q, \;\; \Omega \in \mathcal R^3
\end{align}
where $K$ is the thermal conductivity, $q$ is the internal heat source, and $\boldsymbol{u}$ is the spatially varying temperature field. For multi-task operator learning, we solve this equation on a parametric $3$D circular plate as outlined in \cite{matlab_heattrnsfr}. The plate has an outer diameter of $4$ inches, a central protrusion of height $\frac{1}{4}$ inch, and an overall thickness of $1$ inch. The location and the number of holes are parameterized as follows: the holes are placed at distances $\mathbf{d} \in [0.9, 1.6]$ inches from the center of the protrusion, with the number of holes $\mathbf{n}$ varying between $2$ and $9$. All holes are equispaced in all the considered geometries, while the location of one hole is fixed in all the geometries in the plate (see Figure \ref{fig:3D_plate_params}). The radial location of the holes, $\mathbf{d}$ is varied in steps of $0.1$ inch. Heat sources $\mathbf{q}$ are placed in each hole, with a convective heat flux on the protrusion and side face, and adiabatic conditions elsewhere. These natural convective boundary conditions are shown in Figure \ref{fig:3D_plate_bcs}. The objective is to learn the temperature distribution across the $64$ geometries that result from combinations of $\mathbf{n}$ and $\mathbf{d}$ (details of data generation in Section \ref{sec:data_gen_plate}). Figure \ref{fig:3D_plate_eg} presents four representative geometries and the corresponding temperature fields obtained from MATLAB's PDE Toolbox.

Our objective here is to learn the operator mapping between the geometry parameters defining the different plate configurations and the resulting temperature field due to the heat sources placed inside the holes located in the plate. Since the number and location of the holes change based on the parameters $n$ and $d$, the resulting temperature field varies for each plate configuration. For the operator learning purpose, we select a total of $24$ training samples that encompass the minimum, maximum, and median values of the parameter $d$ for different numbers of holes. For instance, for a plate with $n=3$, we choose three samples with holes located at $d=0.9$, $d=1.2$, and $d=1.6$ as training samples corresponding to $n=3$. The sample choice ensures that the extreme ends of the design space are well represented in the training dataset while avoiding oversampling. The remaining geometries are used as test cases for network inference. A list of the cases used for training and testing is provided in Table~\ref{tab:training_tetsing_samples_heat}.

The parametric representation for the plate ($n$ and $d$) is used as input to our branch network, while the trunk network receives uniformly sampled points in the domain $\Omega = \{(x, y, z) \in \mathbb{R}^3: -2 \leq x \leq 2, -2 \leq y \leq 2, -0.25 \leq z \leq 1\}$ as input. Unlike the previous example on the 2D Darcy problem where the binary mask was considered as input to the branch network, here only the geometry parametrization values are employed. The binary mask is used as additional information in the loss function. We utilize the same grid points for all plate configurations with the trunk network and later apply a binary mask to constrain the temperature field to be zero outside the geometry, similar to the $2$D Darcy problem. The masking matrix is a $3$D grid with values of $0$ (outside) and $1$ (inside) the domain boundary. By utilizing this binary mask, we establish the associativity between the grid points and geometry points, enabling the operator network to learn the temperature field across multiple geometries simultaneously. The training step employs an exponentially decaying learning rate, starting at $1 \times 10^{-3}$ with a decay rate of $0.9$ every $1000$ iterations. 

\begin{figure}[h]
\begin{center}
\includegraphics[width=0.9\textwidth]{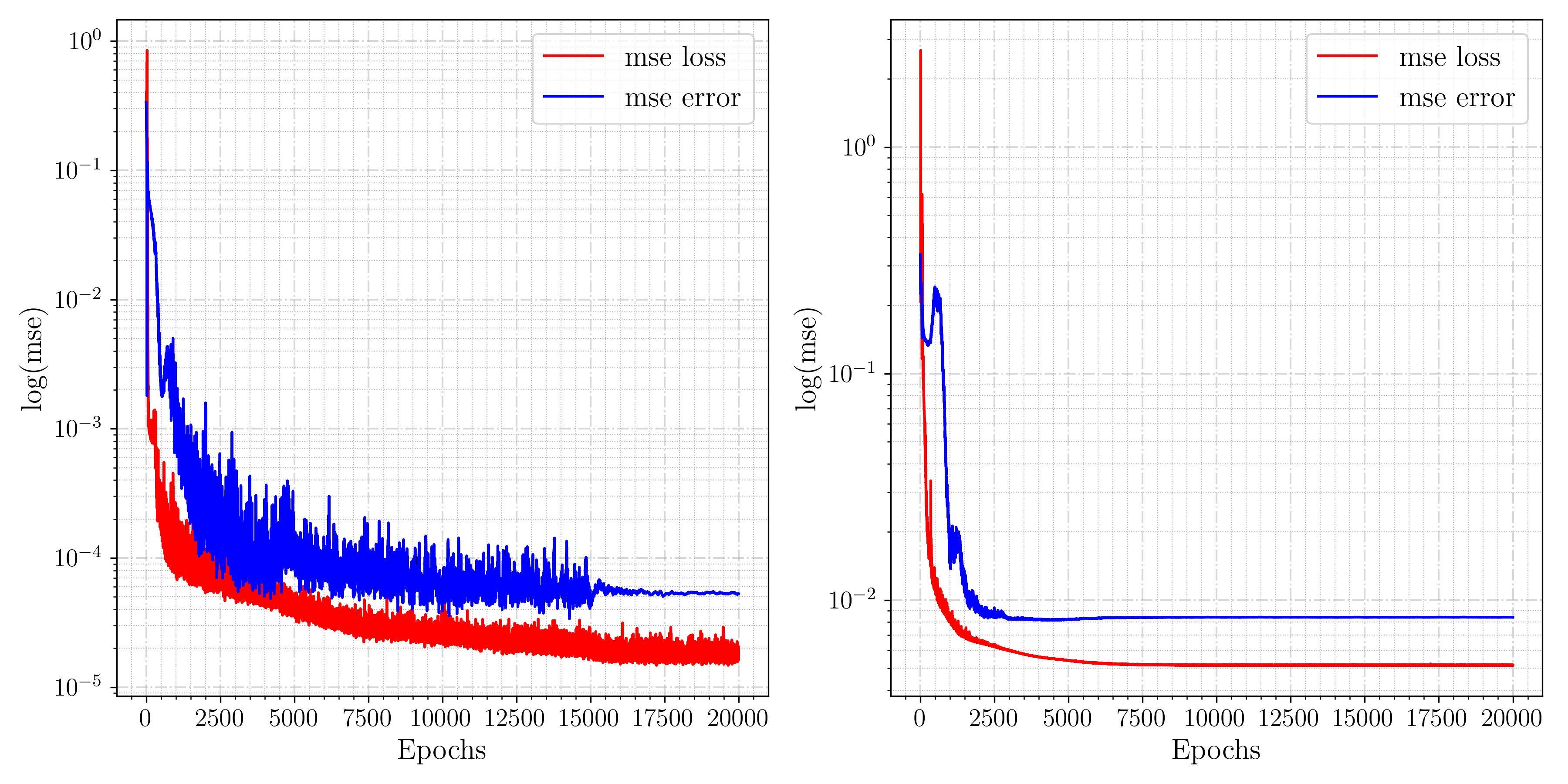}
\caption{Comparison of convergence rates between a network trained with the loss function as a product of the solution operator and the masking function (left) and one trained without a masking operation in the loss function (right). The application of the mask to the solution output demonstrates better error convergence and a reduction in generalization error.}
\label{fig:loss_mask_vs_nomask}
\end{center}
\captionsetup{justification=centering}
\end{figure}

\begin{figure}[h!]
\begin{center}
\includegraphics[width=0.75\linewidth]{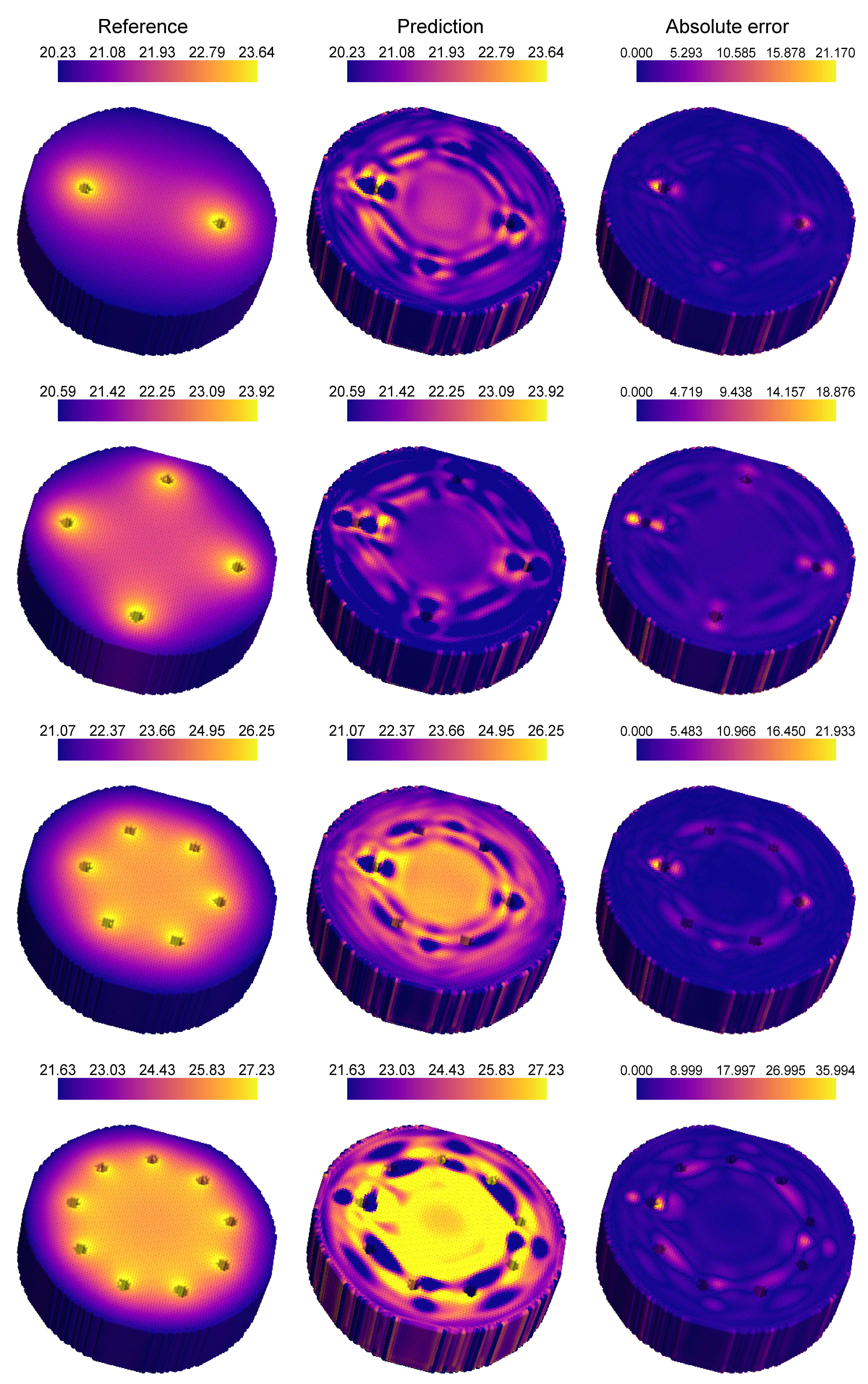}
\caption{Four representative geometries with different numbers and locations of the holes to compare the accuracy of the MT-DeepONet solution (\textbf{without} the masking operation) with the reference solution. The results revealed higher errors across the plate compared to the masked solution (see Figure \ref{fig:heattrnf_solution}). Applying the binary mask imposes an additional constraint that assists the network in learning solutions across varying geometries, particularly when inferring new geometries not seen during training. The masking operation provides crucial geometric information, enhancing the network's ability to generalize to unseen configurations and thereby reducing overall errors.}
\label{fig:mask_vs_nomask_3D}
\end{center}
\captionsetup{justification=centering}
\end{figure}

\begin{figure}[h!]
\begin{center}
\includegraphics[width=0.8\linewidth]{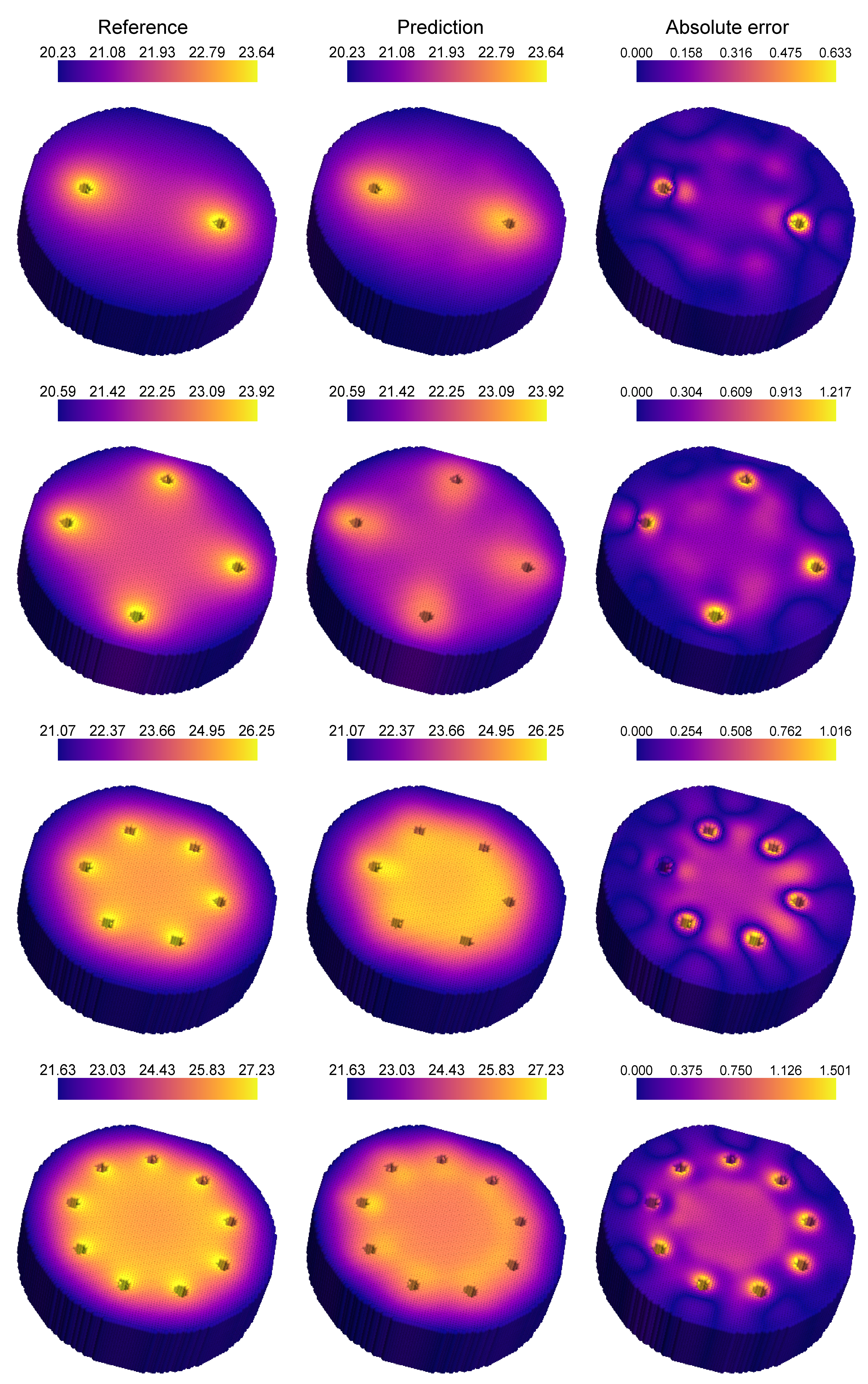}
\caption{Comparison between reference temperature field solution and predictions generated by the MT-DeepONet, for four representative geometries parameterized by number and location of the holes. The prediction shows good agreement with the reference across the domain except for some regions around the holes. We used the parametric representation for the target geometry by providing \{$n,\,p$\} as an input to the branch network for this study and augmented the low-fidelity input data by applying the masking operation on the solution.}
\label{fig:heattrnf_solution}
\end{center}
\captionsetup{justification=centering}
\end{figure}

The binary mask, applied to the network output, ensures that the solution at points outside the geometry is zero, thereby improving convergence and accuracy. Figure \ref{fig:loss_mask_vs_nomask} compares the convergence of the MT-DeepONet with and without the masking function in the loss function. The application of a mask to the solution output demonstrates better error convergence. Figure \ref{fig:mask_vs_nomask_3D} shows the prediction of MT-DeepONet without the masking operation, where higher inaccuracies are observed across the domain due to insufficient geometry representation in the training dataset. Applying a binary mask imposes an extra constraint that assists the network in learning solutions across different geometries. This masking process supplies essential geometric details and improves the network's capacity to generalize to new plate configurations (with the same $n$). Figure \ref{fig:heattrnf_solution} presents representative examples of geometries comparing the reference temperature field with the network prediction. These predictions align well with the reference numerical solution. Higher error values are observed near the holes due to their being under-represented in the training data. We observed better results when more geometries are added to the training dataset but this leads to oversampling in the training dataset. The masking framework functions similarly to transfer learning between tasks, albeit without involving network re-training, and is critical for predicting solutions across different geometries in our experiments. 

\section{Summary}   
\label{sec:summary}

In this work, we introduced the multi-task deep operator network (MT-DeepONet), which learns across multiple scenarios, including different geometries and physical systems, within a single training session. The framework is capable of handling diverse physical processes as demonstrated with the Fisher equation and Darcy flow examples. The MT-DeepONet framework shows a strong ability to map operator solutions across multiple geometries. Thus, the MTL-DeepONet demonstrates significant capabilities in learning and transferring knowledge across varied PDE forms and geometric configurations. While achieving competitive accuracy in predicting solutions, challenges such as managing negative interference between tasks and generalizing to unseen geometries were evident, particularly in the context of learning multiple varied geometries. The binary mask function introduced in the loss term imposes the boundary condition on different geometries as well as improves convergence and generalization. Future research directions should include optimizing network architectures to enhance transfer learning efficacy, developing robust methods for encoding complex $3$D geometries, and exploring advanced techniques to mitigate negative inference in MT-DeepONet.

\clearpage

\section*{Acknowledgements}
The authors would like to acknowledge computing support provided by the Advanced Research Computing at Hopkins (ARCH) core facility at Johns Hopkins University and the Rockfish cluster and the computational resources and services at the Center for Computation and Visualization (CCV), Brown University where all experiments were carried out. 

\section*{Funding}
\noindent VK \& GEK: U.S. Department of Energy project Sea-CROGS (DE-SC0023191) and the OSD/AFOSR Multidisciplinary Research Program of the University Research Initiative (MURI) grant FA9550-20-1-0358.\\
KK: U.S. Department of Energy, Office of Science, Office of Advanced Scientific Computing Research grant under Award Number DE-SC0020428.\\
SG \& MDS: U.S. Department of Energy, Office of Science, Office of Advanced Scientific Computing Research grant under Award Number DE-SC0024162. 

\section*{Author contributions}
\noindent Conceptualization: SG, KK, GEK, MDS \\
Investigation: SG, VK, KK\\
Visualization: VK, SG \\
Supervision: GEK, MDS \\
Writing—original draft: VK, SG\\
Writing—review \& editing: VK, SG, KK, GEK, MDS

\section*{Data and code availability}
\noindent All codes and datasets will be made publicly available at {\small\url{https://github.com/varunsingh88/MTL_DeepONet.git}} upon publication.

\section*{Competing interests}
\noindent Karniadakis has financial interests with the company PredictiveIQ. The rest of the authors declare no competing interests.

\bibliographystyle{elsarticle-num} 
\bibliography{references}

\clearpage
\newpage
\makeatletter
\renewcommand \thesection{S\@arabic\c@section}
\renewcommand\thetable{S\@arabic\c@table}
\renewcommand \thefigure{S\@arabic\c@figure}
\makeatother

\section*{\Large{Supplementary information}}

\setcounter{figure}{0}
\setcounter{table}{0}
\setcounter{section}{0}
\setcounter{page}{1}

\section{Data Generation}
\label{sec:data_generation}
In this section, we present relevant details related to the data generation process for the three problems investigated in this study. 

\subsection{Fisher equations}
For this problem, the goal is to learn the operator mapping from the random initial density condition $u$ across a parameter range $a$ and $b$, as defined in Table \ref{tab:Fisher-KPP}, for its entire time evolution. This mapping is expressed as $\mathcal{G}: u(\boldsymbol{x}, a, b, t=0) \mapsto u(\boldsymbol{x}, t)$, where $t > 0$ and $\boldsymbol{x} \times t \in [0 \times 1] \times [0 \times 1]$. The initial density is modeled as a Gaussian random field (GRF), defined by:
\begin{equation}
\begin{aligned}
\label{eq:GRF_fisher}
    u(\boldsymbol{x}, t=0) \sim \mathcal{GP}(u(\boldsymbol{x}, t=0) \mid \mu(\boldsymbol{x}), \text{Cov}(\boldsymbol{x}, \boldsymbol{x}')),
\end{aligned}
\end{equation}
where $\mu(\boldsymbol{x})$ and $\text{Cov}(\boldsymbol{x}, \boldsymbol{x}')$ are the mean and covariance functions, respectively. We set $\mu(\boldsymbol{x}) = 5 + 0.1 \times \sin(\pi x)$, while the covariance matrix is defined by the squared exponential kernel:
\begin{equation}
\label{eq:A2:cov_fisher}
    \text{Cov}(\boldsymbol{x}, \boldsymbol{x}') = \sigma^2 \exp \left( - \frac{\|x - x'\|_2^2}{\ell^2} \right),
\end{equation}
where $\ell = 0.4$ is the correlation length, and $\sigma^2 = 2$ is the variance. We utilize Karhunen-Lo\'eve expansion (KLE) to generate $1,000$ random initial conditions. The temporal points $t$ and spatial points $\boldsymbol{x}$ are discretized into a $20 \times 64$ grid, resulting in a total of 1,280 collocation points. Parameters $a$ and $b$ are discretized in steps of $0.1$ and $0.15$, respectively, within the defined parameter space. Combining different parameter values for $a$ and $b$ with the random initial conditions, we generate a total of $5,000$ training samples. This dataset is split into training and testing sets using an $80\%-20\%$ split, with $N_{\text{train}} = 4,000$ and $N_{\text{test}} = 1,000$. For preparing data for the multi-task operator network, we concatenate the initial condition data at $64$ domain points with the parameter values for the three Fisher equations defined in Table \ref{tab:Fisher-KPP}, resulting in a $67$ dimensional input vector ($64$ initial condition points + $3$ equation parameters).

\subsection{Darcy equation with transfer learning across multiple geometries} \label{sec:Darcy_datagen}
The multi-task objective in this problem is to learn the operator for the Darcy flow equation described in Equation \ref{eqn:Darcy} over different $2$D spatial domains under randomly generated initial states for conductivity field $K(\boldsymbol{x})$. The hydraulic conductivity field is modeled as a stochastic process, using truncated Karhunen-Lo\'eve expansion. The conductivity field is generated in a reference square domain $\Omega \in [0, 1] \times [0, 1]$ with $100$ grid points using a Gaussian random field with length scale, $\ell = 0.05$. A total of $6000$ solution fields are generated using the multiple initial values of conductivity fields for each geometry. A Dirichlet boundary condition of $h(\boldsymbol{x}) = 0 \quad \forall \boldsymbol{x} \in \partial \Omega $ is applied on all the edges of each geometry, shown in Figure \ref{fig:2D_Darcy_geoms}. The solution to Darcy's equation for all samples is generated using MATLAB's PDE Toolbox solver. The domain mesh is generated using the MATLAB PDE Toolbox mesh generation algorithm with a maximum mesh size of $0.03$ inches for all geometries. We use unstructured triangular elements for the mesh. The hydraulic pressure head solution obtained is linearly interpolated on a regular grid in $\Omega \in [0,1] \times [0, 1]$ to utilize a common trunk network input. Points in the square grid $\Omega$ that lie outside the triangular geometry are extracted and the solution field at such points is manually set to $0$ throughout the dataset. We sub-divide the $6000$ samples into $N_{train} = 5400$ and $N_{test} = 600$ for each source model. For training the MT-DeepONet with multiple geometries, we use an equal number of samples from each geometry to obtain a combined total of $N_{train} = 5400$, thus ensuring equal representation.

\subsection{Heat transfer equation with multiple $3$D geometry} 
\label{sec:data_gen_plate}
To generate data for the steady-state heat transfer problem with multiple plate geometries, we create a parametric CAD model of the plate where the parameters $\boldsymbol{n}$ and $\boldsymbol{d}$ determine the number of holes and their location relative to the central axis of the plate protrusion (see Figure \ref{fig:3D_plate}), respectively. To generate unique geometries with a combination of $n$ and $d$, we fix the location of one hole across all the geometries (see Figure~\ref{fig:3D_plate_params}). We solve the steady-state heat transfer Equation \ref{eqn:heat_transfer} using MATLAB's PDE Toolbox for various design configurations that result from varying the design parameters $\boldsymbol{n}$ and $\boldsymbol{d}$ in the plate geometry, where $2 \leq \boldsymbol{n} \leq 9$ and $0.9 \leq \boldsymbol{d} \leq 1.6$. We increment $\boldsymbol{d}$ in steps of 0.1 resulting in a total of $64$ distinct design configurations. As discussed earlier, we separate the training and test cases based on the hole location, $\boldsymbol{d}$ as shown in Table \ref{tab:training_tetsing_samples_heat}. For each design, we use PDEToolbox for mesh generation with a maximum element size of $0.1$ inch and tetrahedral elements. A heat source $\boldsymbol{q} = 1$ is placed inside each hole, simulating a cartridge heater to heat the plate. Convective boundary conditions are applied on the protruding face and side walls with an ambient temperature $u_\infty = 6$ and a convective heat transfer coefficient $h_c = 0.3$. The output solution, which is the temperature field $u$, is evaluated at all node points of the mesh. Figure \ref{fig:3D_plate_eg} shows representative examples of the different geometries and their corresponding temperature fields obtained by solving Equation \ref{eqn:heat_transfer}. The temperature solutions are initially obtained on unstructured grid points. For simplifying the inputs for the trunk network, we interpolate these solutions over a regular $3$D grid ($100 \times 100\times 100$) around the plate. Points in the regular grid that fall outside the domain of the $3D$ plate are set to $0$ before training the network.

\begin{table}[h]
\centering
\caption{Training and test sample selection for 3D plate problem. We select specific radial locations (d) of the holes for training to encompass the innermost and outermost locations as well as the median location. We interpolate the solution on other radial locations during model inference.}
\label{tab:training_tetsing_samples_heat}
\begin{tabular}{>{\centering\hspace{0pt}}m{0.048\linewidth}>{\centering\hspace{0pt}}m{0.402\linewidth}>{\centering\arraybackslash\hspace{0pt}}m{0.456\linewidth}} 
\toprule
\textbf{n} & \textbf{d}\par{}\textbf{(training samples)} & \textbf{d~}\par{}\textbf{(testing samples)} \\ 
\hline
2 & \{0.9, 1.2, 1.6\} & \{1.0, 1.1, 1.3, 1.4, 1.5\} \\
3 & \{0.9, 1.2, 1.6\} & \{1.0, 1.1, 1.3, 1.4, 1.5\} \\
4 & \{0.9, 1.2, 1.6\} & \{1.0, 1.1, 1.3, 1.4, 1.5\} \\
5 & \{0.9, 1.2, 1.6\} & \{1.0, 1.1, 1.3, 1.4, 1.5\} \\
6 & \{0.9, 1.2, 1.6\} & \{1.0, 1.1, 1.3, 1.4, 1.5\} \\
7 & \{0.9, 1.2, 1.6\} & \{1.0, 1.1, 1.3, 1.4, 1.5\} \\
8 & \{0.9, 1.2, 1.6\} & \{1.0, 1.1, 1.3, 1.4, 1.5\} \\
9 & \{0.9, 1.2, 1.6\} & \{1.0, 1.1, 1.3, 1.4, 1.5\} \\
\bottomrule
\end{tabular}
\end{table}

\section{Data pre-processing}
\label{sec:data_preparation}

The multi-task operator network design for all the four problems discussed earlier consists of a branch and trunk network, where the branch creates a map between the varying input functions (equation representations, geometry, initial conditions) and the output function. The inputs to the branch network are formed by combining the feature representations of the input functions. The input to the trunk network for the Fisher problem consists of spatio-temporal locations $\boldsymbol{x} \times \boldsymbol{t} \in \mathbb R^{n} \times \mathbb R$, where $n = 1, 2$. For the Darcy and $3$D heat transfer problem, the trunk input consists only of spatial points $\boldsymbol{x} \in \mathbb R^{n}$ where $n = 2, 3$. To improve network training, the available data is scaled to reduce to the same order. In particular, the input initial condition and outputs for Darcy, and Fisher problems are scaled using the mean and standard deviation of the training dataset using the operation:
\begin{align}
    x_{in} = \frac{x - \mu}{\sigma}
\end{align}

For the $3$D heat transfer problem, we normalize the inputs and outputs using min-max scaling given by:
\begin{align}
    x_{in} = \frac{x - x_{min}}{x_{max} - x_{min}}
\end{align}
where $x_{max}$ and $x_{min}$ represent the minimum and maximum values across the sample set.

\section{Network Architecture}
\label{sec:network_details}
Our network design and hyper-parameter selection are determined by the problem under investigation. In Table \ref{tab:network_details}, we provide a summary of the network architecture and hyper-parameters used for different problems in this study.  

\begin{table}[h] 
\centering
\caption{Details of network architecture used for different multi-task problems.} \label{tab:network_details}
\resizebox{\linewidth}{!}{%
\begin{tblr}{
  row{1} = {c},
  row{2} = {c},
  cell{1}{1} = {r=2}{},
  cell{1}{2} = {c=2}{},
  cell{1}{4} = {r=2}{},
  cell{1}{5} = {r=2}{},
  cell{1}{6} = {r=2}{},
  cell{1}{7} = {r=2}{},
  cell{1}{8} = {r=2}{},
  cell{3}{2} = {c},
  cell{3}{3} = {c},
  cell{3}{4} = {c},
  cell{3}{5} = {c},
  cell{3}{6} = {c},
  cell{3}{7} = {c},
  cell{3}{8} = {c},
  cell{4}{2} = {c},
  cell{4}{3} = {c},
  cell{4}{4} = {c},
  cell{4}{5} = {c},
  cell{4}{6} = {c},
  cell{4}{7} = {c},
  cell{4}{8} = {c},
  cell{5}{2} = {c},
  cell{5}{3} = {c},
  cell{5}{4} = {c},
  cell{5}{5} = {c},
  cell{5}{6} = {c},
  cell{5}{7} = {c},
  cell{5}{8} = {c},
  cell{6}{2} = {c},
  cell{6}{3} = {c},
  cell{6}{4} = {c},
  cell{6}{5} = {c},
  cell{6}{6} = {c},
  cell{6}{7} = {c},
  cell{6}{8} = {c},
  hline{1,7} = {-}{0.08em},
  hline{2} = {2-3}{},
  hline{3} = {-}{},
}
\textbf{Problem} & \textbf{Branch network} &  & {\textbf{Trunk network}\\\textbf{(neurons per layer)}} & \textbf{Activation} & \textbf{Regularizer} & {\textbf{Dropout}\\\textbf{ (branch)}} & \textbf{Masking}\\
 & \textbf{Type} & \textbf{Neurons per layer} &  &  &  &  & \\
Fisher & MLP & {[}68, 128, 128, 300] & {[}2, 128, 128, 128, 300] & Leaky\_ReLU & $\mathcal{L}_{2}$ & None & None\\
Darcy & {2D CNN\\MLP} & {CNN filters: [16, 32, 64, 64]\\MLP: [128, 128, 150]} & {[}2, 128, 128, 150] & {Conv: Tanh, ReLU\\
    MLP: Leaky\_ReLU} & None & 0.1 & Yes\\
3D heat transfer & MLP & {[}2, 32, 64, 128, 128, 200] & {[}3, 32, 64 $\times$ 3, 128 $\times$ 5, 200] & swish & None & 0.1 & Yes \\ \hline
\end{tblr}}
\end{table}
\end{document}